\documentclass[11pt]{article}

\usepackage[preprint]{acl}

\usepackage{times}
\usepackage{latexsym}

\usepackage[T1]{fontenc}

\usepackage[utf8]{inputenc}

\usepackage{microtype}

\usepackage{inconsolata}

\usepackage{graphicx}
\usepackage{amsmath}
\usepackage{enumitem}
\usepackage{booktabs}
\usepackage{subcaption}
\usepackage{multirow}
\usepackage{multicol}
\usepackage{algorithm}
\usepackage{algorithmic}
\usepackage{colortbl}
\usepackage{ulem}

\definecolor{hlblue}{HTML}{D6EAF8}   
\definecolor{hlgreen}{HTML}{D5F5E3}  

\title{MMR-GRPO: Accelerating GRPO-Style Training through Diversity-Aware Reward Reweighting}

\author{Kangda Wei, Ruihong Huang \\
  Department of Computer Science and Engineering \\Texas A\&M University, College Station, TX\\
  \texttt{\{kangda, huangrh\}@tamu.edu}}

\begin{document}
\maketitle
\begin{abstract}
Group Relative Policy Optimization (GRPO) \cite{shao2024deepseekmathpushinglimitsmathematical} has become a standard approach for training mathematical reasoning models; however, GRPO training is computationally intensive and usually takes a long time, which consumes substantial computational resources and creates barriers for academic researchers and smaller organizations with limited GPU budgets. 
In this paper, we propose MMR-GRPO to accelerate GRPO training and reduce the overall training time required to reach peak performance, and the approach adopts Maximal Marginal Relevance to reweigh rewards of multiple rollouts by balancing rollout quality with diversity to reduce rollout redundancy. 
The rationale is that redundant or similar rollout, when repeatedly used to train a model, will create an ``exploitation trap'' and slow down model convergence in GRPO style reinforcement learning. 
Extensive evaluations across three model sizes (1.5B, 7B, 8B), 
three GRPO variants, 
and five mathematical reasoning benchmarks show that MMR-GRPO achieves comparable peak performance while requiring on average 47.9\% fewer training steps and 70.2\% less wall-clock time. These gains are consistent across models, methods, and benchmarks. Our code is released at: \url{https://github.com/WeiKangda/MMR-GRPO}.
\end{abstract}

\section{Introduction}

\begin{figure}[t]
    \centering
    \includegraphics[width=\linewidth]{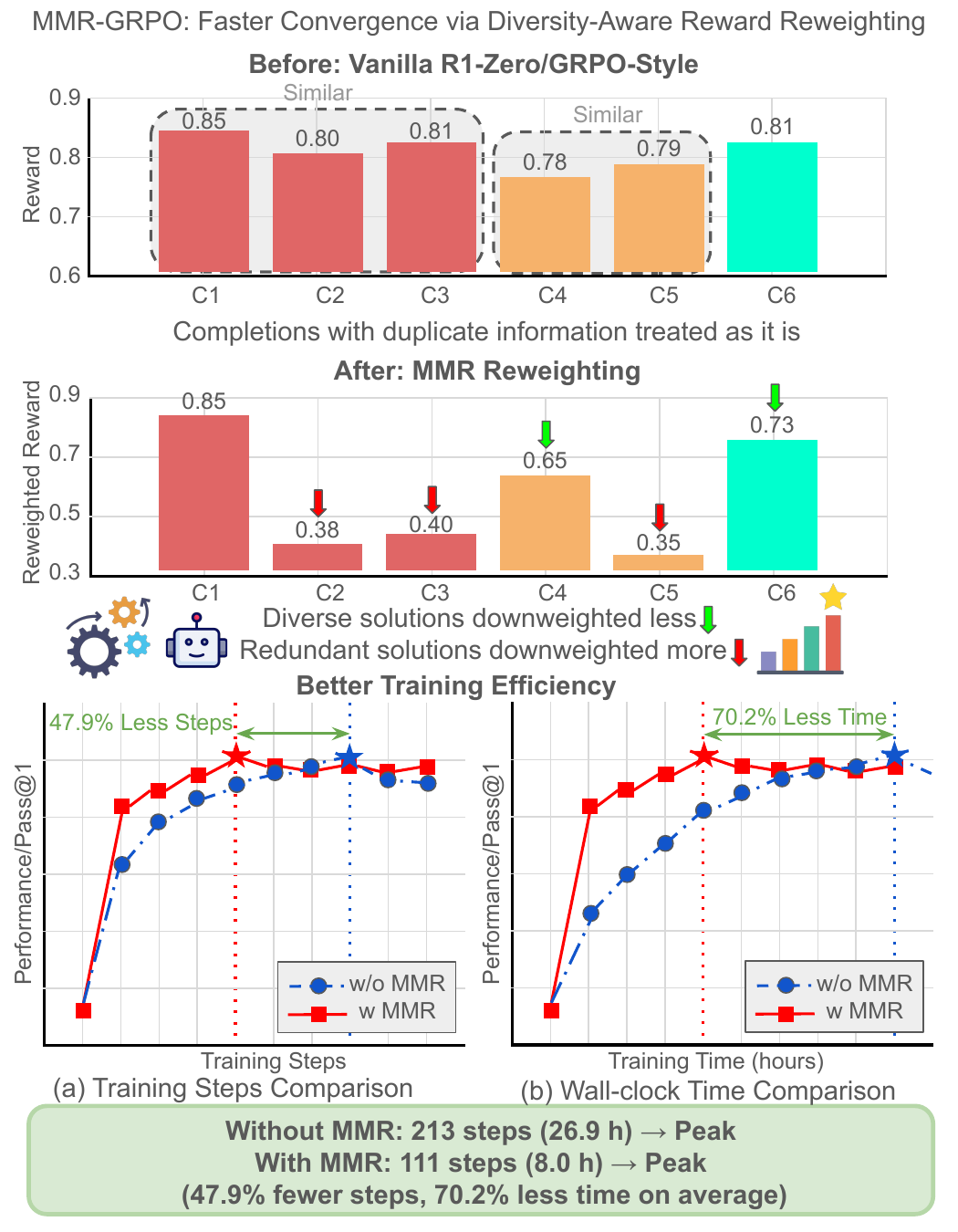}
    \vspace{-0.3cm}
    \caption{Before MMR, all completions receive similar rewards despite semantic redundancy (C1-C3 are similar, C4-C5 are similar). After MMR reweighting, diverse completions (C1, C4, C6) maintain high rewards while redundant ones (C2, C3, C5) are downweighted. Training with MMR achieves comparable peak performance with 47.9\% fewer training steps and 70.2\% less wall-clock time on average.}
    \label{fig:intro}
    \vspace{-0.2cm}
\end{figure}

Recent advances in large language models (LLMs) have demonstrated remarkable capabilities in mathematical reasoning tasks \cite{wei2023chainofthoughtpromptingelicitsreasoning, wang-etal-2024-math}, with reinforcement learning (RL) emerging as a critical technique for aligning models with complex reasoning objectives \cite{ahn-etal-2024-large, wang2025surveylargelanguagemodels, trung-etal-2024-reft, kazemnejad2025vinepporefiningcreditassignment, setlur2024rewardingprogressscalingautomated, gehring2025rlefgroundingcodellms, li2024policy}. Group Relative Policy Optimization (GRPO) or R1-Zero-style training \cite{shao2024deepseekmathpushinglimitsmathematical, deepseekai2025deepseekr1incentivizingreasoningcapability} has become the de facto standard for training state-of-the-art mathematical reasoning models \cite{srivastava2025technicalsurveyreinforcementlearning}, achieving strong performance across diverse benchmarks requiring multi-step problem-solving and logical inference. 

However, GRPO-style reinforcement learning (RL) usually takes many steps and a long time to reach peak performance. 
Further, the training process is highly computationally intensive as GRPO-style RL relies on generating multiple completions per prompt and requires frequent model inference throughout training \cite{nimmaturi2025predictivescalinglawsefficient}. 
As a consequence, GRPO training 
consumes substantial energy and computational resources, with training cost and carbon footprints growing proportionally to training duration \cite{patterson2021carbonemissionslargeneural, faiz2024llmcarbonmodelingendtoendcarbon}. Moreover, lengthy training time creates barriers for academic researchers and smaller organizations with limited GPU budgets, concentrating advanced reasoning capabilities among well-resourced institutions \cite{besiroglu2024computedividemachinelearning, ahmed2020dedemocratizationaideeplearning, Gelles_Kinoshita_Musser_Dunham_2024}. Therefore, methods that achieve comparable performance with reduced training time will reduce environmental impact and democratize access to state-of-the-art reasoning models \cite{khandelwal2025100k100daystradeoffs}.

In this paper, we propose \textbf{MMR-GRPO} to promote exploration in GRPO-style RL and accelerate model convergence, where ``MMR'' refers to ``Maximal Marginal Relevance'', a classic technique used for re-ranking search results to reduce redundancy. 
Similarly, MMR-GRPO will 
reweight rewards of completions in each group by balancing completion quality with diversity to reduce redundancy among completions. 

The rationale is that redundant or similar completions, if repeatedly used to train a model, will slow down model convergence in GRPO style RL. As a model is trained on its own generated completions in GRPO style RL, 
if a model keeps generating the same completion or highly overlapped completions, 
this shows that such completions align well with its current policy, 
then, training repeatedly with such highly redundant completions will create an ``exploitation trap'' and overly bias the model toward its current policy, 
limiting exploration of RL and slowing model convergence.
Therefore, we strive to reduce redundancy within each group of completions to prevent ``exploitation trap'', 
balance exploitation and exploration,  
and achieve faster convergence. 
A similar effect is observed in deep RL, where diversity-based experience replay accelerates learning over redundant sampling~\cite{zhao2025efficientdiversitybasedexperiencereplay}.

As illustrated by Figure \ref{fig:intro}, MMR-GRPO reweighs rewards during advantage computation by penalizing completions that are semantically similar to higher-rewarded completions already being selected. More concretely, we compute sentence embeddings for all completions and apply a greedy MMR selection procedure that iteratively selects the next completion 
with the highest adjusted reward, adjusted based on its maximal semantic similarity with the already selected completions.


Further, we introduce a parameter-free adaptive mechanism where the diversity-quality trade-off $\lambda$ is automatically determined by the standard deviation of completion rewards within each group: when rewards are tightly concentrated, diversity is prioritized to encourage exploration; when rewards are spread, quality is emphasized. This eliminates manual hyperparameter tuning while maintaining effectiveness across different model scales and training regimes.

We evaluate MMR-GRPO across three model sizes (1.5B, 7B, 8B parameters), three GRPO variants (the original GRPO, DR-GRPO and DAPO), and five mathematical reasoning benchmarks (MATH-500, AIME 2024, AMC 2023, Minerva Math, OlympiadBench). Our results demonstrate consistent and substantial training efficiency gains. MMR-GRPO achieves comparable performance as baseline methods while requiring 47.9\% fewer training steps on average to reach peak performance. Despite MMR adding only 1-5\% per-step computational overhead due to embedding computation and similarity matrix operations, the substantial reduction in total training steps translates to 70.2\% wall-clock time savings overall. 

In summary, our contributions include:
\begin{enumerate}[noitemsep,topsep=0pt]
    \item We propose \textbf{MMR-GRPO}, a diversity-aware reward reweighting approach that reduces both the number of training steps and wall-clock training time required to reach peak performance, while maintaining comparable performance. The method can be seamlessly integrated into GRPO–style training paradigms.
    \item We introduce a parameter-free adaptive mechanism that balances rollout quality and diversity without hyperparameter tuning.
    \item We provide comprehensive evaluation across three model scales, three methods, and five benchmarks, showing consistent efficiency gains.
\end{enumerate}

\section{Related Work}
\subsection{Reinforcement Learning for Mathematical Reasoning in LLMs}
Reinforcement learning has emerged as a critical technique for aligning language models with complex reasoning objectives. Modern RL methods for LLMs evolved from foundational algorithms—Vanilla Policy Gradient \cite{Williams2004SimpleSG}, TRPO \cite{schulman2017trustregionpolicyoptimization}, and PPO \cite{schulman2017proximalpolicyoptimizationalgorithms}—into specialized variants for language model training, such as GRPO \cite{shao2024deepseekmathpushinglimitsmathematical} and later DAPO \cite{yu2025dapoopensourcellmreinforcement}. 
For mathematical reasoning specifically, GRPO \cite{shao2024deepseekmathpushinglimitsmathematical}, or R1-Zero-style training \cite{deepseekai2025deepseekr1incentivizingreasoningcapability}, has become the de facto standard. Recent extensions of GRPO prioritizes algorithmic improvements for better performance \cite{li2025discoreinforcinglargereasoning, chen2025dragrpoexploringdiversityawarereward, nan2025ngrponegativeenhancedgrouprelative, bamba2025xrpopushinglimitsgrpo} over training efficiency—the focus of our approach.

DAPO \cite{yu2025dapoopensourcellmreinforcement} also proposes dynamic sampling, a technique that repeatedly generates candidate samples and discards low-variance sample sets until a high-variance sample set is obtained. 
Dynamic sampling was proposed to improve training efficiency and reduce the total number of training steps needed for model convergence. However, even though dynamic sampling reduces training steps by 50-67\%, recent analysis reveals a critical limitation: each dynamic sampling step requires approximately 1.5\textasciitilde3 times longer wall-clock time \cite{li2025discoreinforcinglargereasoning, lian2025comparativeanalysisparametrictuning} due to multiple rounds of generation and filtering \cite{nemo_rl_dapo}.
Consequently, even though dynamic sampling reduces training steps, DAPO's actual wall-clock training time remains comparable to or even exceeds vanilla GRPO \cite{chen2025lspolengthawaredynamicsampling}, limiting its practical efficiency gains. 

\subsection{Low-Variance Sample Groups in GRPO} 
In GRPO-style training, each update is based on a group of sampled completions, with policy gradients computed from the relative rewards within the group. However, when the sampled group exhibits low reward variance—for instance, when all completions are similarly incorrect or similarly correct—the resulting advantages become weak or uninformative, leading to inefficient policy updates and slower convergence.
Prior work has addressed this issue through special handling or avoidance of low-variance groups. 
DAPO \cite{yu2025dapoopensourcellmreinforcement} and Xrpo \cite{bamba2025xrpopushinglimitsgrpo} simply discard low-variance sample sets, reducing the number of training steps required to reach peak performance. CPPO \cite{lin2025cppo} accelerates GRPO by pruning completions with low absolute advantages to reduce forward passes during policy updates, though it operates on sample selection and does not modify how retained completions are rewarded, which is orthogonal to our approach. Other methods modify the learning signal to better exploit low-variance or negative samples, by separating the loss computation for positive and negative samples \cite{zhu2025surprisingeffectivenessnegativereinforcement}, or introducing a virtual perfect-score reward to reshape advantages when all rewards are zero \cite{nan2025ngrponegativeenhancedgrouprelative}.

In contrast, our method directly exploits low-variance groups rather than discarding or treating them differently. By uniformly reweighing samples based on semantic diversity for any group of samples, we extract informative learning signals even when the reward variance of a group is low. As a result, our approach accelerates learning and not only reduces the number of training steps required for convergence, but also achieves consistent wall-clock time savings.

\subsection{Diversity-Aware Methods for GRPO}
Several concurrent works also incorporate diversity signals into GRPO training. DIVER \cite{hu2025diversityincentivizedexplorationversatilereasoning} introduces an intrinsic diversity reward via potential-based reward shaping, but requires maintaining a potential function across steps and heuristics to mitigate reward hacking. DARLING \cite{li2025jointlyreinforcingdiversityquality} trains a semantic equivalence classifier to cluster completions for diversity scoring, requiring domain-specific annotations. 

DRA-GRPO \cite{chen2025dragrpoexploringdiversityawarereward} 
 reweighs the rewards of all the completions using Submodular Mutual Information (SMI) solely based on pairwise cosine similarities or levels of redundancy. 
MMR-GRPO shares DRA-GRPO's lightweight design but differs in the reweighting mechanism by considering both completion redundancy and completion quality.
After reward reweighting of DRA-GRPO, there may not be any high quality completion retaining a high reward if the high quality completions all have high similairities with each other or the other completions in the group, on the other hand, a low quality completion may gain a relatively high reward just because the completion is unique. In contrast, MMR-GRPO reduces redundancy in a group of completions by balancing completion quality with completion diversity. 
In our comparison experiments (Section~\ref{sec:dra_comparison}) adopting the robust multi-sampling approach ($n>1$) for calculating the metric pass@1, our approach MMR-GRPO consistently outperforms DRA-GRPO in maintaining the peak performance and in significantly reducing training steps. 

\section{Method}
\subsection{Background: GRPO}
GRPO \cite{shao2024deepseekmathpushinglimitsmathematical} is an effective reinforcement learning method for aligning language models with human preferences. Unlike traditional policy gradient methods, GRPO optimizes the policy by generating multiple responses 
and computing advantages within a group of responses.

Given a prompt $x$, GRPO generates $G$ completions $\{y_1, y_2, \ldots, y_G\}$ and computes a reward $r(y_i)$ for each completion using a reward model. The advantage for each completion is calculated as:
\begin{equation}
\label{eq:advantage}
A(y_i) = \frac{r(y_i) - \mu_G}{\sigma_G + \epsilon}
\end{equation}
where $\mu_G$ and $\sigma_G$ are the mean and standard deviation of rewards within the group, and $\epsilon$ is a small constant for numerical stability.

The training objective combines the policy gradient with a KL divergence penalty to prevent the policy from deviating too far from a reference model:
\begin{equation}
\label{eq:loss}
\scalebox{0.9}{$
\mathcal{L} = -{E}_{y \sim \pi_\theta}\left[\log \pi_\theta(y|x) \cdot A(y) - \beta \cdot D_{\text{KL}}(\pi_\theta || \pi_{\text{ref}})\right]$}
\end{equation}
where $\pi_\theta$ is the trainable policy, $\pi_{\text{ref}}$ is the reference policy, and $\beta$ is the KL penalty coefficient.

While GRPO effectively leverages group-relative rewards, it does not explicitly account for diversity among the generated completions. High-reward responses that are semantically similar may dominate the training signal, leading to slower convergence.

\subsection{MMR-based Reward Reweighting with $\lambda$}

To address the lack of diversity consideration in vanilla GRPO, we propose a MMR inspired reward reweighting mechanism. MMR \cite{10.1145/290941.291025} was originally designed for document retrieval to balance relevance and diversity. We adapt this principle to reward shaping in GRPO. The complete algorithm is shown in Algorithm~\ref{alg:mmr_reweighting}.

Let $\mathcal{E}(y_i) \in {R}^d$ denote the embedding of completion $y_i$, obtained using a pre-trained sentence encoder. We define the cosine similarity between two completions as $m(y_i, y_j) = \mathcal{E}(y_i)^\top \mathcal{E}(y_j)$.

The MMR-based reward adjustment follows a greedy selection procedure. We maintain a set $\mathcal{S}$ of selected completions and iteratively add completions that maximize a diversity-weighted score:
\begin{equation}
\label{eq: mmr}
\text{score}(y_i) = \lambda \cdot r(y_i) - (1 - \lambda) \cdot \max_{y_j \in \mathcal{S}} m(y_i, y_j)
\end{equation}
where $\lambda \in [0, 1]$ is a hyperparameter controlling the trade-off between reward quality (relevance) and diversity, and is commonly set to 0.7. When $\lambda = 1$, the method reduces to standard reward-based selection; when $\lambda = 0$, it purely maximizes diversity.

The algorithm proceeds as follows:
\begin{enumerate}[noitemsep,topsep=2pt,parsep=0pt]
    \item Initialize $\mathcal{S} = \emptyset$ and precompute the similarity matrix $M \in {R}^{G \times G}$ where $M_{ij} = m(y_i, y_j)$. 
    \item Select the completion with the highest reward: $y^* = \arg\max_{y_i} r(y_i)$, and add it to $\mathcal{S}$.
    \item For each remaining completion, compute its adjusted score based on Equation~\ref{eq: mmr}.
    \item Select the completion with the highest score, add it to $\mathcal{S}$.
    \item Repeat steps 3 and 4 until all completions are ranked.
\end{enumerate}
Finally, return the adjusted scores as reweighted rewards $\tilde{r}(y_i)$, which will replace the original rewards $r(y_i)$ when computing advantages  (Equation~\ref{eq:advantage}). 

Note that after MMR reweighting, semantically similar correct answers will receive different advantages, which can be potentially confusing to the model. But we explain below that this is fine and preferred. 
The advantage assigned to a rollout will essentially become a multiplier on the token-level training loss, and this multiplier intuitively corresponds to the attention we would like the model to pay during training. For semantically similar correct answers, we prioritize learning from the first occurrence and subsequent repetitions should receive less attention to avoid over-training on an already familiar rollout, which could slow down exploration and learning. Meanwhile, low-quality answers will not receive a high advantage simply for being distinct as a low reward $r(y_i)$ will be assigned initially and 
the diversity term only modulates the original reward, 
according to Equation~\ref{eq: mmr}.

\begin{algorithm}[t]
\caption{MMR-based Reward Reweighting}
\label{alg:mmr_reweighting}
\small
\begin{algorithmic}[1]
\REQUIRE Rewards $\{r(y_1), \ldots, r(y_G)\}$, L2-normalized embeddings $\{\mathcal{E}(y_1), \ldots, \mathcal{E}(y_G)\}$
\ENSURE Reweighted rewards $\{\tilde{r}(y_1), \ldots, \tilde{r}(y_G)\}$
\STATE Compute adaptive $\lambda$: $\lambda_{\text{adapt}} \gets \sigma(\text{std}(r)) = \frac{1}{1 + e^{-\text{std}(r)}}$
\STATE Compute similarity matrix: $M_{ij} \gets \mathcal{E}(y_i)^\top \mathcal{E}(y_j)$ for all $i, j \in [G]$ at once.
\STATE Initialize $\mathcal{S} \gets \emptyset$ 
\STATE $i^* \gets \arg\max_{i} r(y_i)$
\STATE Add $i^*$ to $\mathcal{S}$ and set $\tilde{r}(y_{i^*}) \gets r(y_{i^*})$
\FOR{$t = 1$ to $G-1$}
    \FOR{each $i \notin \mathcal{S}$}
        \STATE $\text{best\_sim}_i \gets \max_{j \in \mathcal{S}} M_{ij}\ $
    \ENDFOR
    \FOR{each $i \notin \mathcal{S}$}
        \STATE $\text{score}(y_i) \gets \lambda_{\text{adapt}} \cdot r(y_i) - (1 - \lambda_{\text{adapt}}) \cdot \text{best\_sim}_i$
    \ENDFOR
    \STATE $i^* \gets \arg\max_{i \notin \mathcal{S}} \text{score}(y_i)$
    \STATE Add $i^*$ to $\mathcal{S}$ and set $\tilde{r}(y_{i^*}) \gets \text{score}(y_{i^*})$
\ENDFOR
\STATE \textbf{return} $\{\tilde{r}(y_1), \ldots, \tilde{r}(y_G)\}$
\end{algorithmic}
\end{algorithm}
\vspace{-0.2cm}

\subsection{Parameter-free MMR Reweighting with Adaptive $\lambda$}

While the $\lambda$-parameterized MMR reweighting is effective, it introduces an additional hyperparameter that requires tuning. To eliminate this dependency, we design an adaptive mechanism that automatically adjusts $\lambda$ based on the distribution of rewards within each group.

The key insight is that when rewards have high variance, diversity is less critical because the model already explores different quality regions. Conversely, when rewards are similar, diversity becomes more important to avoid mode collapse. We formalize this intuition using the sigmoid function applied to the reward standard deviation:
\begin{equation}
\label{eq: sigmoid}
\scalebox{0.9}{$
\lambda_{\text{adapt}} = \sigma(\text{std}(r)) = \frac{1}{1 + e^{-\text{std}(r)}}$}
\end{equation}
where $\text{std}(r)$ denotes the standard deviation of rewards $\{r(y_1), \ldots, r(y_G)\}$ within a group.


This adaptive $\lambda$ is \textbf{scale-invariant} as the sigmoid function automatically maps reward variability to a bounded range $(0.5, 1)$, making it robust to different reward scales across tasks. 
Empirically, when rewards are tightly concentrated (low $\text{std}(r)$), $\lambda_{\text{adapt}}$ decreases toward 0.5, increasing the diversity penalty. When rewards are widely spread (high $\text{std}(r)$), $\lambda_{\text{adapt}}$ approaches 1, prioritizing reward quality. This adaptive behavior provides a principled way to encourage diversity without sacrificing reward maximization when the model already exhibits sufficient exploration.

The parameter-free MMR reweighting follows the same greedy procedure as described in Section 3.2, but with $\lambda$ replaced by $\lambda_{\text{adapt}}$ computed automatically for each group of completions. This allows the model to dynamically balance relevance and diversity based on the reward landscape of each group. 
All the experiments in this paper are conducted with $\lambda_{\text{adapt}}$,  unless noted otherwise.

\section{Experimental Settings}
\subsection{Datasets}
We evaluate our approach on five widely used mathematical reasoning benchmarks: AIME 2024, MATH 500 \cite{hendrycks2021measuringmathematicalproblemsolving}, AMC 2023, Minerva \cite{lewkowycz2022solvingquantitativereasoningproblems}, and Olympiad Bench \cite{he-etal-2024-olympiadbench}.
These benchmarks span competition-style and curriculum-aligned problems with varying levels of difficulty and reasoning complexity, providing a comprehensive evaluation to assess mathematical problem-solving and multi-step reasoning ability. Additional details about each benchmark are provided in Appendix~\ref{app:testing_data}.

For training, we use the \texttt{knoveleng/open-rs} dataset \cite{dang2025reinforcementlearningreasoningsmall}, which consists of mathematical reasoning problems paired with high-quality step-by-step solutions.
The dataset covers a broad range of mathematical topics and difficulty levels and is well-suited for training models to generate coherent reasoning chains. Further details about the training data are described in Appendix~\ref{app:training_data}.

\subsection{Evaluation Metrics}

Following standard practice in evaluation from previous works \cite{deepseekai2025deepseekr1incentivizingreasoningcapability, yu2025dapoopensourcellmreinforcement, li2025discoreinforcinglargereasoning}, we adopt the \textbf{pass@k} metric to measure the probability that at least one correct solution exists among $k$ generated candidates from $n$ total samples. Formally, given $n=16$ independently generated completions per problem, pass@k is computed as:
\vspace{-0.2cm}
\begin{equation}
\text{pass@k} = {E} \left[ 1 - \frac{\binom{n-c}{k}}{\binom{n}{k}} \right]
\end{equation}
where $c$ is the number of correct solutions among the $n$ samples. We primarily report \textbf{pass@1} with $n=16$. When $k=1$, this metric is equivalent to \textbf{avg@16}, which is the fraction of correct solutions among the 16 sampled completions.. Evaluations are done using \textit{lighteval} \footnote{\tiny \url{https://huggingface.co/docs/lighteval}} for reproducibility and fair comparison.

\subsection{Models}

We conduct experiments with three model scales from two different model families of Deepseek-AI\footnote{\tiny \url{https://huggingface.co/deepseek-ai}}: DeepSeek-R1-Distill-Qwen-1.5B
, DeepSeek-R1-Distill-Qwen-7B, and DeepSeek-R1-Distill-Llama-8B. All models are initialized from publicly available checkpoints from Huggingface. Larger models (7B and 8B) employ LoRA \cite{hu2021loralowrankadaptationlarge} for parameter-efficient fine-tuning (see Appendix~\ref{app:lora} for details). For embedding extraction, we use \texttt{jina-embeddings-v2-small-en}\footnote{\tiny \url{https://huggingface.co/jinaai/jina-embeddings-v2-small-en}}, which provide strong semantic representations while maintaining computational efficiency.

\newcommand{\cb}[1]{\cellcolor{hlblue}#1}   
\newcommand{\cg}[1]{\cellcolor{hlgreen}#1}  
\begin{table*}[t]
\centering
\small
\setlength{\tabcolsep}{4pt}
\scalebox{0.75}{
\begin{tabular}{@{}l|l|cccccc|ccc@{}}
\toprule
\textbf{Size} & \textbf{Method/Model} & \textbf{AIME 24} & \textbf{MATH-500}  & \textbf{AMC 23} & \textbf{Minerva} & \textbf{OlympiadBench} & \textbf{Average} & \begin{tabular}[c]{@{}c@{}}\textbf{Peak}\\\textbf{Step}\end{tabular} & \begin{tabular}[c]{@{}c@{}}\textbf{Time}\\\textbf{(hrs)}\end{tabular} & \begin{tabular}[c]{@{}c@{}}\textbf{Time per}\\\textbf{Step(s)}\end{tabular}\\
\midrule
\multirow{8}{*}{\textbf{1.5B}}
& DS-Distill-Qwen-1.5B & 0.288 & 0.828 & 0.629 & 0.265 & 0.433 & 0.489 & - & - & - \\
\cmidrule(lr){2-11}
& GRPO & \cg{0.338} & 0.846 & 0.730 & 0.296 & 0.528 & 0.547 & 100 & \cb{4.08} & \cb{147} \\
& MMR-GRPO & 0.325 & \cb{0.849} & \cb{0.739} & \cb{0.302} & 0.528 & \cb{0.549} & 100 & 4.13 & 149 \\
\cmidrule(lr){2-11}
& DR-GRPO & \cg{0.335} & 0.844 & \cb{0.744} & 0.297 & 0.523 & 0.549 & 150 & 6.11 & \cb{147} \\
& MMR-DR-GRPO & 0.323 & \cb{0.851} & 0.738 & \cb{0.303} & \cb{0.530} & 0.549 & \cg{100} & \cg{4.13} & 149 \\
\cmidrule(lr){2-11}
& DAPO & 0.331 & 0.851 & 0.755 & 0.304 & 0.541 & 0.556 & 110 & 25.53 & 836 \\
& DAPO-No-DS & \cg{0.348} & 0.855 & \cg{0.744} & \cb{0.298} & \cb{0.529} & \cb{0.555} & 200 & 8.93 & \cb{161} \\
& MMR-DAPO-No-DS & 0.331 & \cb{0.856} & 0.730 & 0.295 & 0.527 & 0.548 & \cb{170} & \cb{7.72} & 163 \\
\midrule
\midrule
\multirow{8}{*}{\textbf{7B}}
& DS-Distill-Qwen-7B & 0.560 & 0.923 & 0.825 & 0.380 & 0.568 & 0.651 & - & - & - \\
\cmidrule(lr){2-11}
& GRPO & 0.554 & 0.940 & \cb{0.917} & \cb{0.418} & 0.671 & 0.700 & 350 & 28.28 & \cb{291} \\
& MMR-GRPO & \cb{0.560} & 0.940 & 0.916 & 0.409 & \cb{0.673} & 0.700 & \cg{150} & \cg{12.22} & 293 \\
\cmidrule(lr){2-11}
& DR-GRPO & 0.565 & 0.939 & \cb{0.914} & \cb{0.420} & 0.672 & \cb{0.702} & 300 & 24.30 & \cb{292} \\
& MMR-DR-GRPO & 0.565 & \cb{0.942} & 0.905 & 0.412 & \cb{0.673} & 0.699 & \cg{50} & \cg{4.11} & 296 \\
\cmidrule(lr){2-11}
& DAPO & 0.558 & 0.940 & 0.914 & 0.418 & 0.671 & 0.700 & 90 & 33.15 & 1326 \\
& DAPO-No-DS & \cb{0.569} & 0.939 & 0.905 & \cb{0.420} & \cb{0.674} & 0.701 & 200 & 16.17 & \cb{291} \\
& MMR-DAPO-No-DS & 0.567 & \cb{0.941} & \cg{0.920} & 0.417 & 0.672 & \cb{0.703} & \cg{100} & \cg{8.52} & 307 \\
\midrule
\midrule
\multirow{8}{*}{\textbf{8B}}
& DS-Distill-Llama-8B & 0.506 & 0.896 & 0.815 & 0.295 & 0.541 & 0.611 & - & - & - \\
\cmidrule(lr){2-11}
& GRPO & 0.465 & \cb{0.889} & 0.897 & \cb{0.355} & \cb{0.626} & \cb{0.646} & 350 & 32.22 & \cb{331} \\
& MMR-GRPO & \cg{0.475} & 0.882 & 0.897 & 0.350 & 0.623 & 0.645 & \cg{50} & \cg{4.62} & 333 \\
\cmidrule(lr){2-11}
& DR-GRPO & \cb{0.488} & \cb{0.895} & 0.881 & \cb{0.351} & \cb{0.632} & \cb{0.649} & 300 & 27.72 & \cb{333} \\
& MMR-DR-GRPO & 0.485 & 0.893 & \cb{0.886} & 0.346 & 0.630 & 0.648 & \cg{100} & \cg{9.36} & 337 \\
\cmidrule(lr){2-11}
& DAPO & 0.504 & 0.889 & 0.889 & 0.351 & 0.631 & 0.653 & 160 & 93.75 & 2109 \\
& DAPO-No-DS & \cb{0.483} & \cb{0.890} & 0.878 & 0.346 & 0.628 & 0.645 & 250 & 23.02 & \cb{331} \\
& MMR-DAPO-No-DS & 0.477 & 0.888 & 0.878 & \cb{0.353} & \cb{0.631} & \cb{0.646} & \cg{180} & \cb{17.40} & 348 \\
\midrule
\midrule
\multicolumn{3}{r}{\textbf{Average Training Step Saved:}}
& \multicolumn{2}{l|}{\textbf{47.9\%}}
& \multicolumn{4}{r}{\textbf{Average Training Time (hrs) Saved:}}
& \multicolumn{2}{l}{\textbf{70.2\%}} \\
\bottomrule
\end{tabular}}
\vspace{-0.1cm}
\caption{Peak performance (pass@1, $n=16$) comparisons across model sizes (1.5B, 7B and 8B models) and training methods (GRPO, DR-GRPO and DAPO), before and after MMR reweighting. All metrics represent the best checkpoint for each configuration. The Peak Step indicates the training step where optimal average performance is achieved. Training time is logged based on wall-clock measurements on 2$\times$NVIDIA H100 80GB GPUs. For MMR v.s. Non-MMR methods, \colorbox{hlblue}{blue} = better but comparable; \colorbox{hlgreen}{green} = significantly better ($\geq$1\% for performance, $>$25\% for efficiency). Unshaded = equivalent. MMR reweighting consistently achieves comparable or better performance while requiring fewer training steps and less time, translating to substantial computational savings. DS is short for DeepSeek-R1.}
\vspace{-0.4cm}
\label{tab:peak_performance}
\end{table*}

\subsection{Training Methods}
We experiment with three GRPO-style reinforcement learning methods for aligning language models with mathematical reasoning objectives:

\begin{itemize}[noitemsep,topsep=1pt,parsep=0pt,leftmargin=0.5em]
    \item \textbf{GRPO} \cite{shao2024deepseekmathpushinglimitsmathematical} applies PPO-style policy gradient optimization \cite{schulman2017proximalpolicyoptimizationalgorithms} by generating multiple completions per prompt and computes advantages relative to group statistics, avoiding the need for a separate value network. We compare vanilla GRPO and MMR-GRPO, with both methods trained for 500 steps and evaluated every 50 steps.
    \item \textbf{DAPO} \cite{yu2025dapoopensourcellmreinforcement} introduces dynamic sampling, a technique that reduces training steps by discarding low-variance sample sets and regenerating samples during training. However, dynamic sampling incurs significant per-step computational overhead \cite{li2025discoreinforcinglargereasoning, lian2025comparativeanalysisparametrictuning}. To investigate whether MMR can serve as a more efficient alternative to dynamic sampling, we evaluate three DAPO configurations:
        \begin{itemize}[noitemsep,topsep=0pt,parsep=0pt,leftmargin=1.0em]
            \item \textbf{DAPO}: Vanilla DAPO with dynamic sampling enabled, trained for 200 steps with evaluation after every 10 steps. The reduced training horizon is due to dynamic sampling's rapid convergence as well as its prohibitive per-step cost.
            \item \textbf{DAPO-No-DS}: DAPO with dynamic sampling disabled, providing a controlled baseline without dynamic sampling. Trained for 500 steps with evaluation first after every 10 steps (0-200) and then after every 50 steps (200-500).
            \item \textbf{MMR-DAPO-No-DS}: DAPO with MMR reweighting instead of dynamic sampling, using the same training and evaluation schedule as DAPO-No-DS. In this configuration, we do not discard any low-variance sample group, but we consistently reweigh rewards within each group of generated samples.  
        \end{itemize}
    This experimental design allows us to directly compare MMR and Dynamic Sampling as alternative training efficiency techniques while isolating their effects from DAPO's other algorithmic improvements.
    \item \textbf{DR-GRPO} \cite{liu2025understandingr1zeroliketrainingcritical} extends GRPO with de-biased optimization that reduce response-level length bias and question-level difficulty bias. We compare vanilla DR-GRPO and MMR-DR-GRPO, with both methods trained for 500 steps and evaluated every 50 steps.
\end{itemize}

Common training hyperparameters and other model details are provided in Appendix~\ref{app:training}.

\begin{figure*}[t]
    \centering
    \begin{subfigure}[t]{\textwidth}
        \centering
        \includegraphics[width=\linewidth]{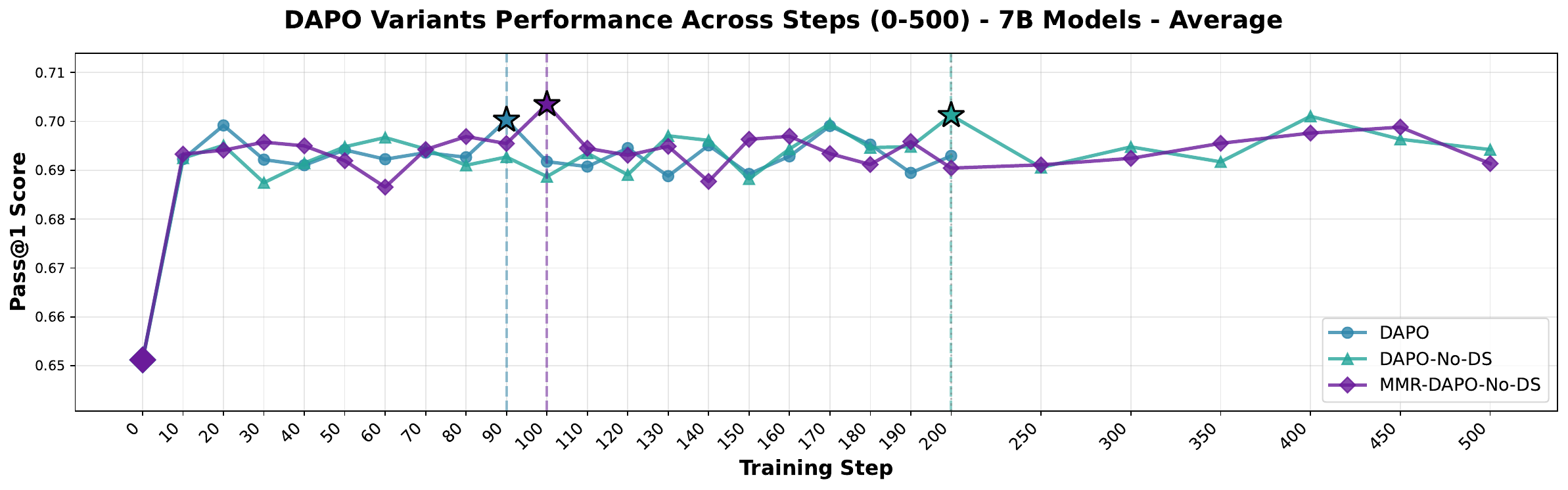}
    \end{subfigure}

    \begin{subfigure}[t]{0.48\textwidth}
        \centering
        \includegraphics[width=\linewidth]{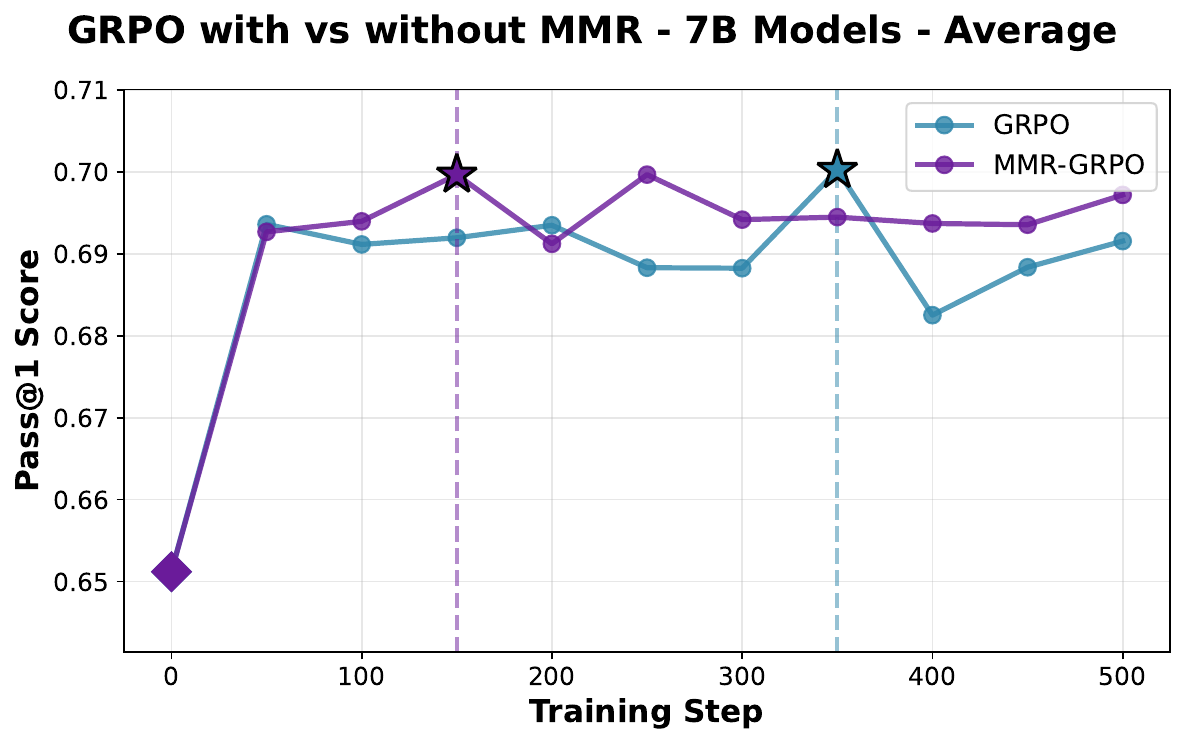}
    \end{subfigure}
    \hfill
    \begin{subfigure}[t]{0.48\textwidth}
        \centering
        \includegraphics[width=\linewidth]{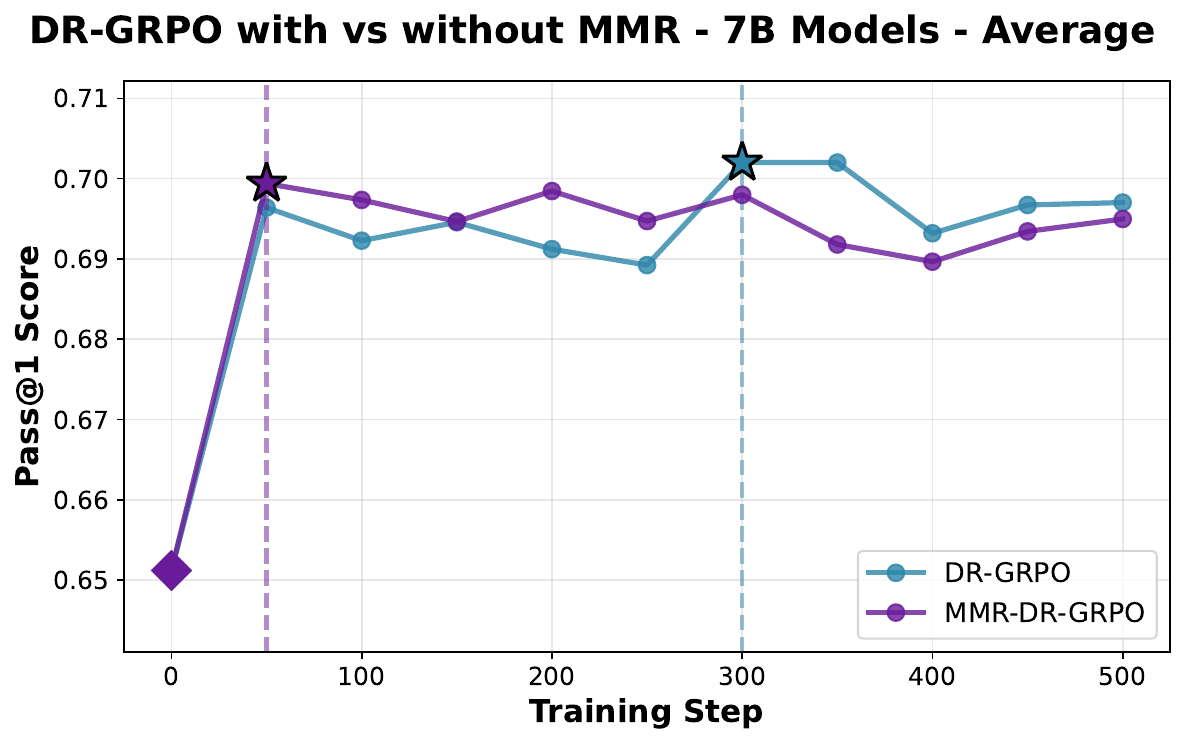}
    \end{subfigure}
    \vspace{-0.3cm}
    \caption{Performance across training steps for 7B models across all three training methods (DAPO, GRPO, DR-GRPO). MMR variants consistently achieve faster convergence and reach peak performance with fewer training steps.}
    \vspace{-0.3cm}
    \label{fig:7B}
\end{figure*}

\section{Results and Discussions}
\subsection{MMR Impacts on Training Efficiency}
\paragraph{Training Steps Reduction}
Table~\ref{tab:peak_performance} demonstrates that MMR-based reward reweighting consistently reduces the number of training steps required to reach peak performance across all methods and model scales. For GRPO, MMR achieves comparable performance while reducing training steps from 350 to 150 steps for the 7B model (57\% reduction) and from 350 to 50 steps for the 8B model (86\% reduction), though the 1.5B model shows no reduction in training steps (both methods peak at 100 steps), which is mainly due to its smaller model size, as smaller models typically require fewer steps to converge. Similarly, for DR-GRPO, MMR-DR-GRPO reduces training steps from 150 to 100 steps (33\% reduction) for 1.5B, from 300 to 50 steps (83\% reduction) for 7B, and from 300 to 100 steps (67\% reduction) for 8B models. For DAPO without dynamic sampling, MMR reduces steps from 200 to 170 (15\% reduction) for 1.5B, from 200 to 100 (50\% reduction) for 7B, and from 250 to 180 (28\% reduction) for 8B models. On average, MMR achieves a 47.9\% reduction in training steps while maintaining comparable peak performance. Figure~\ref{fig:7B} illustrates this faster convergence for 7B models: MMR variants reach their peak performance earlier than their baseline counterparts, and results for 8B and 1.5B models across all benchmarks (Figures~\ref{fig:8B} and~\ref{fig:1.5B}) exhibit similar patterns (detailed in Appendix~\ref{app:convergence_7b_8b}).


\paragraph{Wall-clock Training Time Savings}
Beyond step reduction, MMR delivers substantial wall-clock training time savings, addressing a critical limitation of existing efficiency techniques. Table~\ref{tab:peak_performance} reveals that DAPO with dynamic sampling, despite achieving peak performance in fewer nominal steps (90-160 steps), incurs drastically longer wall-clock times: 25.53 hours for 1.5B, 33.15 hours for 7B, and 93.75 hours for 8B models. This counterintuitive inefficiency stems from dynamic sampling's multi-round generation and filtering mechanism, which increases per-step computational cost by 5$\times$ compared to DAPO-No-DS, DAPO with no dynamic sampling. 
In contrast, MMR-DAPO-No-DS achieves similar peak performance while requiring only 7.72, 8.52, and 17.40 hours respectively—representing 70\%, 74\%, and 81\% wall-clock time savings over vanilla DAPO. For GRPO and DR-GRPO, MMR similarly delivers dramatic time reductions: MMR-GRPO reduces training time from 28.28 to 12.22 hours (57\% savings) for 7B and from 32.22 to 4.62 hours (86\% savings) for 8B models, while MMR-DR-GRPO achieves even more impressive gains, reducing time from 24.30 to 4.11 hours (83\% savings) for 7B and from 27.72 to 9.36 hours (66\% savings) for 8B models. Across all configurations, MMR achieves an average of 70.2\% wall-clock time savings—a substantial practical improvement that translates to lower computational costs, reduced carbon emissions, and improved accessibility for researchers with limited GPU budgets.
\begin{figure*}[t]
    \centering
    \begin{subfigure}{0.32\linewidth}
        \centering
        \includegraphics[width=\linewidth]{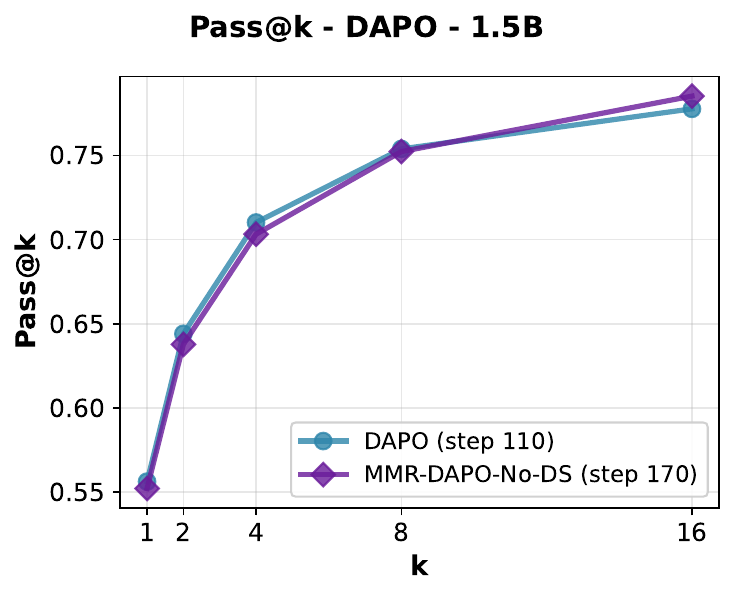}
    \end{subfigure}
    \hfill
    \begin{subfigure}{0.32\linewidth}
        \centering
        \includegraphics[width=\linewidth]{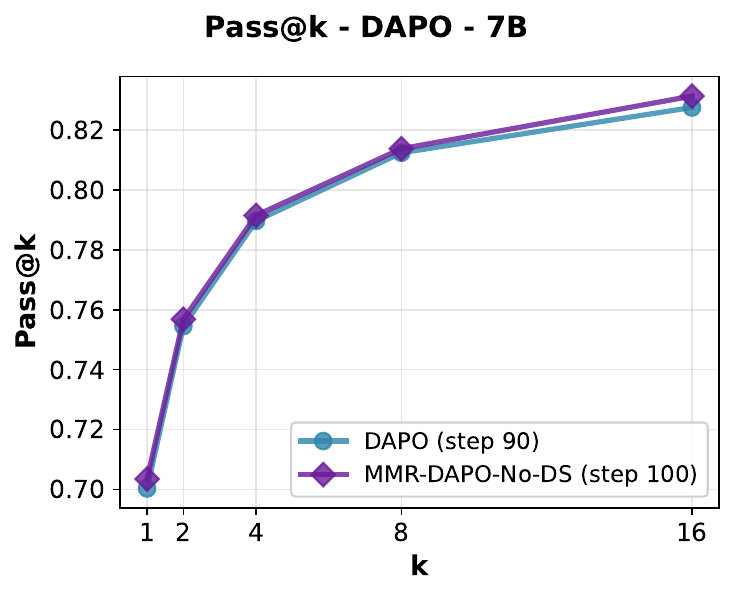}
    \end{subfigure}
    \hfill
    \begin{subfigure}{0.32\linewidth}
        \centering
        \includegraphics[width=\linewidth]{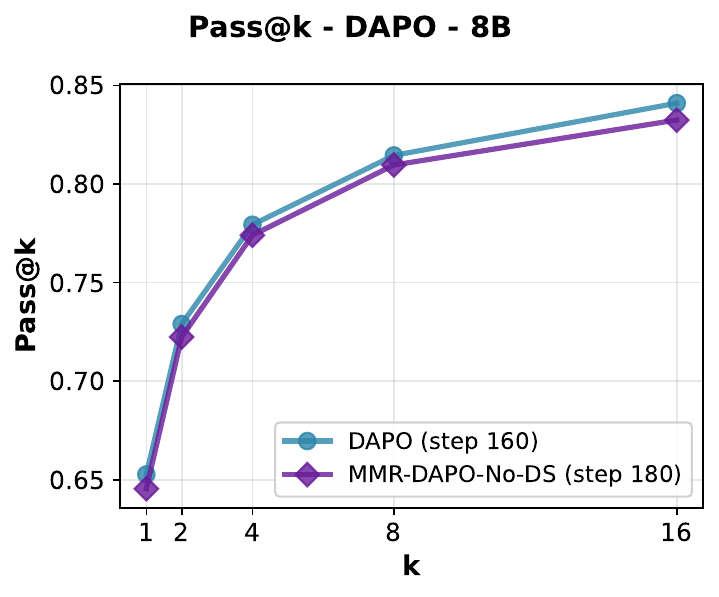}
    \end{subfigure}
    \vspace{-0.4cm}
    \caption{Pass@k curves for DAPO methods across three model scales (1.5B, 7B, 8B). MMR-DAPO-No-DS maintains nearly identical pass@k performance to vanilla DAPO across all k values (1, 2, 4, 8, 16), demonstrating that MMR training does not harm exploration capability.}
    \vspace{-0.2cm}
    \label{fig:pass_at_k_dapo}
\end{figure*}

\paragraph{Minimal Per-step Computational Overhead}
A critical consideration for any training efficiency method is the per-step computational overhead it introduces. Table~\ref{tab:peak_performance} shows that MMR adds minimal per-step cost: for 1.5B models, time per step increases from 147s to 149s for GRPO (+1.4\%), from 147s to 149s for DR-GRPO (+1.4\%), and from 161s to 163s for DAPO-No-DS (+1.2\%). Similar patterns hold for 7B models and 8B models. 
This 1-5\% overhead is negligible compared to the computational demands of model forward/backward passes and generation, making this modest per-step cost more than offset by the substantial reduction in total training steps.
The contrast with DAPO's dynamic sampling is striking: while MMR adds 1-5\% per step, dynamic sampling increases per-step time by 412\% (836s vs 163s at 1.5B), 331\% (1326s vs 307s at 7B), and 506\% (2109s vs 348s at 8B), consistent with prvious works finding \cite{li2025discoreinforcinglargereasoning, lian2025comparativeanalysisparametrictuning}. This demonstrates that MMR achieves training efficiency gains through accelerated convergence rather than expensive per-step filtering. 

\subsection{Comparisons with DRA-GRPO}
\label{sec:dra_comparison}
We further compare MMR-GRPO against DRA-GRPO~\cite{chen2025dragrpoexploringdiversityawarereward}, another recent diversity-aware GRPO variant that reweighs completions based on semantic diversity. Table~\ref{tab:mmr_vs_dra} reports peak avg@16 (pass@1, $n=16$) and the corresponding peak training step for vanilla GRPO, DRA-GRPO, and MMR-GRPO across 1.5B, 7B, and 8B models. Under the standard multi-sampling evaluation\footnote{It appears to us that DRA-GRPO~\cite{chen2025dragrpoexploringdiversityawarereward} reported pass@1 results use k=1 and n = 1 (i.e., a single sample), which measures raw accuracy rather than the standard average@k metric with a larger sample size (e.g., k=16), which we and the broader community adopt.}, DRA-GRPO 
consistently underperforms vanilla GRPO by a small amount. Regarding peak training steps, while DRA-GRPO has 
reduced the number of training steps on the 8B model, DRA-GRPO takes 100 more training steps on the 7B model to reach peak performance. 
In contrast, MMR-GRPO achieves comparable avg@16 and requires substantially fewer training steps, indicating that explicitly penalizing redundancy via MMR provides a stronger and more efficient training signal than DRA-GRPO.

\begin{table}[t]
\centering
\small
\setlength{\tabcolsep}{4pt}
\begin{tabular}{@{}l|cc|cc|cc@{}}
\toprule
\multirow{2}{*}{\textbf{Method}} & \multicolumn{2}{c|}{\textbf{1.5B}} & \multicolumn{2}{c|}{\textbf{7B}} & \multicolumn{2}{c}{\textbf{8B}} \\
& \textbf{Avg.} & \textbf{Step} & \textbf{Avg.} & \textbf{Step} & \textbf{Avg.} & \textbf{Step} \\
\midrule
GRPO       & 0.547 & 100 & 0.700 & 350 & \textbf{0.646} & 350 \\
DRA-GRPO   & 0.545 & 100 & 0.699 & 450 & 0.644 & 150 \\
MMR-GRPO   & \textbf{0.549} & \textbf{100} & \textbf{0.700} & \textbf{150} & 0.645 & \textbf{50} \\
\bottomrule
\end{tabular}
\vspace{-0.2cm}
\caption{Peak avg@16 (pass@1, $n=16$) and peak training step for vanilla GRPO, DRA-GRPO, and MMR-GRPO across 1.5B, 7B, and 8B models.}
\vspace{-0.4cm}
\label{tab:mmr_vs_dra}
\end{table}

\subsection{MMR Effects on Trained Model Exploration Ability}
A potential concern with using MMR is whether it might harm the model's ability to explore diverse solution strategies at inference time. Figure~\ref{fig:pass_at_k_dapo} addresses this concern by comparing pass@k curves (k=1 to 16) for DAPO and MMR-DAPO-No-DS across all three model scales. The curves are nearly overlapping across all k values, indicating that MMR-trained models maintain comparable exploration ability to their baseline counterparts. For instance, at k=16, all three methods achieve similar coverage (approximately 0.78 for 1.5B, 0.83 for 7B, and 0.83 for 8B models), suggesting that diversity-aware reward reweighting during training does not constrain the model's capacity to generate varied solutions during inference. This is intuitive: while MMR downweighs semantically redundant completions during training to accelerate learning, it does not restrict the model from learning novel reasoning paths. 
The trained model retains the ability to explore multiple reasoning paths, as evidenced by the comparable pass@k performance. Similar patterns hold for GRPO and DR-GRPO methods (Figures~\ref{fig:pass_at_k_grpo} and~\ref{fig:pass_at_k_drgrpo} in Appendix~\ref{app:passk_grpo_drgrpo}). These results confirm that reweighting the completions rewards by using MMR to penalize completions with duplicate information does not restrict the model's exploration capabilities, making it a practical method for reducing training costs without sacrificing inference-time solution diversity.

\subsection{Ablation Study}
\begin{table}[t]
\centering
\small
\begin{tabular}{@{}lcccc@{}}
\toprule
\multirow{2}{*}{\textbf{Strategy}} & \multicolumn{2}{c}{\textbf{MMR-GRPO}} & \multicolumn{2}{c}{\textbf{MMR-DR-GRPO}} \\
\cmidrule(lr){2-3} \cmidrule(lr){4-5}
& \textbf{Pass@1} & \textbf{Steps} & \textbf{Pass@1} & \textbf{Steps} \\
\midrule
Fixed $\lambda$=0.5 & 0.5495 & 150 & 0.5579 & 200 \\
Fixed $\lambda$=0.6 & 0.5546 & 150 & 0.5515 & 50 \\
Fixed $\lambda$=0.7 & 0.5548 & 200 & \textbf{0.5605} & \textbf{150} \\
Fixed $\lambda$=0.8 & 0.5499 & 50 & 0.5470 & 50 \\
Fixed $\lambda$=0.9 & \textbf{0.5636} & \textbf{100} & 0.5497 & 50 \\
\midrule
\textbf{Adaptive $\lambda$} & \textbf{0.5583} & \textbf{100} & \textbf{0.5589} & \textbf{100} \\
\bottomrule
\end{tabular}
\vspace{-0.2cm}
\caption{Ablation study on the diversity-relevance trade-off parameter $\lambda$ for 1.5B models. Pass@1 scores are computed with n=1 due to budget restriction.}
\vspace{-0.4cm}
\label{tab:lambda_ablation}
\end{table}
\paragraph{Adaptive $\lambda$ Mechanism:} Table~\ref{tab:lambda_ablation} compares our adaptive $\lambda$ mechanism against fixed values on 1.5B models. A key finding is that \textbf{the optimal fixed $\lambda$ differs across methods}: $\lambda$=0.9 performs best for MMR-GRPO while $\lambda$=0.7 is optimal for MMR-DR-GRPO. This method-dependent sensitivity would require costly hyperparameter search for each new training algorithm. In contrast, adaptive $\lambda$, requiring no hyperparameter tuning, achieves competitive performance for both methods after a relatively small number of training step, further demonstrating robustness and the practical value of automatic adaptation. 

\paragraph{Robustness to Embedding Model:} A natural question is whether our method is sensitive to the choice of embedding model used for computing semantic similarity. To investigate this, we compare our default embedding model, \texttt{jina-embeddings-v2-small-en} against \texttt{nomic-embed-text-v1.5} on MMR-GRPO using 7B model, keeping all other hyperparameters identical. As shown in Table~\ref{tab:embedding_ablation}, the two models yield nearly indistinguishable results: the average benchmark score differs by only 0.1\%, peak performance is reached at the same training step, and wall-clock time differs by less than 1\%. Per-benchmark scores are equally consistent, with a maximum deviation of 0.5\% on AIME 2024. These results confirm that our method is robust to the choice of embedding model, as the MMR reweighting mechanism relies on \textit{relative} similarity rankings among completions rather than absolute embedding magnitudes, making it insensitive to embedding model-specific representation differences.

\begin{table}[t]
\centering
\small
\begin{tabular}{lccc}
\toprule
& \textbf{nomic} & \textbf{jina} & \textbf{$\Delta$} \\
\midrule
AIME 2024 & 0.565 & 0.560 & 0.005 \\
MATH-500 & 0.940 & 0.940 & 0.000 \\
AMC 2023 & 0.914 & 0.916 & -0.002 \\
Minerva & 0.412 & 0.409 & 0.003 \\
OlympiadBench & 0.673 & 0.673 & 0.000 \\
\midrule
Average & 0.701 & 0.700 & 0.001 \\
Peak Step & 150 & 150 & 0 \\
Time (hrs) & 12.30 & 12.22 & 0.08 \\
\bottomrule
\end{tabular}
\vspace{-0.2cm}
\caption{Embedding model robustness ablation on MMR-GRPO. nomic = nomic-embed-text-v1.5; jina = jina-embeddings-v2-small-en.}
\vspace{-0.4cm}
\label{tab:embedding_ablation}
\end{table}

\section{Conclusions}
We propose MMR-GRPO, a diversity-aware reward reweighting method for accelerating model learning in GRPO-style training. By downweighting semantically redundant completions using Maximal Marginal Relevance, MMR-GRPO provides more informative policy updates and accelerates convergence. Across three model sizes, three GRPO variants, and five benchmarks, MMR-GRPO matches peak performance while reducing training steps by 47.9\% and wall-clock time by 70.2\%. 
Overall, MMR-GRPO lowers training cost while preserving performance, democratizing access to advanced reasoning model development and reduces the environmental impact of RL training.

\section*{Limitations}
While our proposed MMR-based reward reweighting achieves comparable performance to baseline methods with significantly fewer training steps and less training time, we acknowledge several limitations. 

First, the greedy MMR selection algorithm has quadratic time complexity $\mathcal{O}(N^2)$, where $N$ is the number of generated completions per prompt, compared to $\mathcal{O}(N)$ for standard GRPO reward normalization. However, this overhead is minimal in practice: our experiments use $N=6$ generations per prompt (standard in prior works \cite{chen2025dragrpoexploringdiversityawarereward, dang2025reinforcementlearningreasoningsmall}), the MMR reweighting adds only 1 - 5\% to training time per step since model generation and forward passes remain the dominant computational costs. Our fully vectorized GPU implementation leverages efficient batched matrix multiplication, even with $N=64$  ($64^2 \times 512 = 2.1$M FLOPs for $N=64$ with 512-dimensional embeddings), making the similarity matrix computation negligible compared to a single forward pass through a multi-billion parameter language model. For context, a single forward pass through a 1.5B parameter model requires approximately $3 \times 10^9$ FLOPs (assuming 2 FLOPs per parameter). The MMR computation at N=64 represents only 0.07\% of a single forward pass. Moreover, the similarity matrix is computed once per training step using fully vectorized GPU operations (batched matrix multiplication), adding negligible wall-clock overhead. Even scaling to N=128 or N=256 would remain practically insignificant compared to the model's generation and gradient computation costs. Importantly, the computational overhead of MMR is more than offset by the reduced training steps required to reach peak performance: our method achieves optimal results in 47.9\% fewer steps on average, resulting in substantial wall-clock time savings despite the per-step overhead. 

Second, due to computational constraints, our experiments are limited to models up to 8B parameters using LoRA fine-tuning for larger scales. We have not evaluated our approach on larger models (14B, 32B, or 70B parameters) with full fine-tuning. Given that our method's efficiency gains stem from encouraging diverse exploration early in training, we hypothesize that larger models with greater capacity for memorizing diverse solution strategies may benefit even more from MMR reweighting, though empirical validation remains future work. 

Finally, our evaluation focuses on mathematical reasoning benchmarks; the generalization of MMR-based reward reweighting to other domains such as code generation or commonsense reasoning, which may exhibit different convergence dynamics and solution space characteristics, remains to be explored.

\section*{Ethics and Broder Impacts}
This work improves the training efficiency of reinforcement learning for mathematical reasoning models, achieving comparable performance with 47.9\% fewer training steps and 70.2\% less training time. This reduction in computational requirements has positive environmental implications by lowering energy consumption and carbon emissions, and democratizes access to advanced reasoning model development for researchers with limited budgets. However, we acknowledge that more efficient training of capable reasoning models could accelerate deployment in high-stakes domains such as educational assessment and automated theorem proving, which require careful validation and human oversight to prevent misuse (e.g., automated exam cheating). Our evaluation is limited to English-language mathematical benchmarks; equitable access to reasoning AI requires extending such methods to multilingual settings. We encourage practitioners to implement appropriate safeguards and emphasize that responsible AI development requires investments in robustness testing, bias mitigation, and value alignment beyond computational efficiency.

\section*{Acknowledgments}

We would like to thank the anonymous reviewers for their valuable feedback and input. We gratefully acknowledge support from National Science Foundation via the award IIS-1942918. Portions of this research were conducted with the advanced computing resources provided by Texas A\&M High-Performance Research Computing.

\bibliography{anthology,custom}

\appendix
\section{Dataset Details}
\label{app:datasets}

This section provides detailed information about the benchmarks used for evaluation and the training dataset.

\subsection{Evaluation Benchmarks}
\label{app:testing_data}
We evaluate our approach on five mathematical reasoning benchmarks that cover diverse problem types, difficulty levels, and mathematical domains. Table~\ref{tab:dataset_stats} summarizes the key statistics of each benchmark.

\begin{table}[h]
\centering
\small
\scalebox{0.9}{\begin{tabular}{lcc}
\toprule
\textbf{Benchmark} & \textbf{\# Problems} & \textbf{Difficulty} \\
\midrule
MATH-500 & 500 & High School \\
AIME 2024 & 30 & Competition \\
AMC 2023 & 40 & High School Competition \\
Minerva Math & 272 & Undergraduate \\
OlympiadBench & 675 & Competition \\
\bottomrule
\end{tabular}}
\vspace{-0.2cm}
\caption{Statistics of evaluation benchmarks. Difficulty levels indicate the target audience or competition level.}
\label{tab:dataset_stats}
\end{table}

\paragraph{MATH-500} \cite{hendrycks2021measuringmathematicalproblemsolving}
A curated subset of 500 challenging problems from the MATH dataset, spanning seven mathematical domains: Prealgebra, Algebra, Number Theory, Counting and Probability, Geometry, Intermediate Algebra, and Precalculus.

\paragraph{AIME 2024} 
The American Invitational Mathematics Examination (AIME) 2024 consists of 30 problems designed for top-performing high school students who qualify through the AMC competitions. All answers are integers between 0 and 999, requiring precise numerical solutions. Problems demand multi-step reasoning, creative problem-solving strategies, and deep mathematical insight, making this one of the most challenging benchmarks in our evaluation suite.

\paragraph{AMC 2023}
The American Mathematics Competitions (AMC) 2023 includes 40 problems from AMC 10 and AMC 12 contests. These problems cover fundamental to intermediate mathematical concepts including algebra, geometry, number theory, and combinatorics. Compared to AIME, AMC problems are more accessible but still require solid mathematical reasoning and problem-solving skills. Answers are typically integers or simple algebraic expressions.

\paragraph{Minerva Math} \cite{lewkowycz2022solvingquantitativereasoningproblems}
Originally curated to evaluate the Minerva model, this benchmark contains 272 undergraduate-level mathematics problems. Problems often involve complex multi-step derivations, symbolic manipulation, and application of advanced mathematical concepts.

\paragraph{OlympiadBench} \cite{he-etal-2024-olympiadbench}
A comprehensive collection of 675 competition-level mathematics problems sourced from international and national mathematical olympiads (IMO, USAMO, etc.). These problems represent the pinnacle of pre-collegiate mathematical problem-solving, often requiring creative insights, elegant proofs, and sophisticated mathematical techniques.

\subsection{Training Dataset}
\label{app:training_data}

We use the \texttt{knoveleng/open-rs} dataset \cite{dang2025reinforcementlearningreasoningsmall} for training all models. This dataset is specifically designed for reinforcement learning on mathematical reasoning tasks and contains high-quality step-by-step solutions following a structured format.

\paragraph{Dataset Composition}
The dataset consists of 1.7k mathematical problems paired with detailed reasoning chains. Each example includes:
\begin{itemize}[noitemsep,topsep=1pt,parsep=0pt,leftmargin=1.5em]
    \item \textbf{Problem}: A natural language mathematical question
    \item \textbf{Reasoning process}: Step-by-step solution with the final answer in \texttt{\textbackslash\textbackslash boxed\{\}}
    \item \textbf{Final answer}: The value of the final answer
\end{itemize}

\section{Training Hyperparameters}
\label{app:training}

This section provides detailed training configurations for all methods and model scales used in our experiments.

\subsection{Model Configurations}
\label{app:lora}
Table~\ref{tab:model_configs} summarizes the architecture and parameter-efficient fine-tuning configurations for all three model scales.

\begin{table*}[h]
\centering
\scalebox{0.8}{
\begin{tabular}{lccc}
\toprule
\textbf{Configuration} & \textbf{1.5B} & \textbf{7B} & \textbf{8B} \\
\midrule
Model name & \multicolumn{2}{c}{DeepSeek-R1-Distill-Qwen} & DeepSeek-R1-Distill-Llama \\
Architecture & Qwen & Qwen & Llama \\
Parameters & 1.5B & 7B & 8B \\
Fine-tuning method & Full FT & LoRA & LoRA \\
\midrule
LoRA rank ($r$) & -- & 64 & 64 \\
LoRA alpha ($\alpha$) & -- & 128 & 128 \\
LoRA dropout & -- & 0.05 & 0.05 \\
LoRA target modules & -- & \multicolumn{2}{c}{\small q/k/v/o/gate/up/down proj} \\
\midrule
Precision & \multicolumn{3}{c}{bfloat16} \\
Attention & \multicolumn{3}{c}{Flash Attention 2} \\
\bottomrule
\end{tabular}}
\caption{Model architecture and PEFT configurations. All models are initialized from publicly available DeepSeek-R1 distilled checkpoints.}
\label{tab:model_configs}
\end{table*}

\subsection{Common Training Hyperparameters}

Table~\ref{tab:common_hyperparams} lists the training hyperparameters shared across all methods and model scales.

\begin{table}[h]
\centering
\small
\scalebox{0.8}{
\begin{tabular}{lc}
\toprule
\textbf{Hyperparameter} & \textbf{Value} \\
\midrule
\multicolumn{2}{l}{\textit{Optimization}} \\
Learning rate & $1.0 \times 10^{-6}$ \\
Optimizer & AdamW \\
LR scheduler & Cosine with min lr \\
Min LR ratio & 0.1 \\
Warmup ratio & 0.1 \\
Gradient clipping & 1.0 \\
\midrule
\multicolumn{2}{l}{\textit{Batch Configuration}} \\
Batch size per device & 6 \\
Gradient accumulation steps & 8 \\
Effective batch size & 48 \\
\midrule
\multicolumn{2}{l}{\textit{Training Steps}} \\
Max steps (GRPO/DR-GRPO/DAPO w/o DS) & 500 \\
Max steps (DAPO) & 200 \\
Logging steps & 1 \\
\midrule
\multicolumn{2}{l}{\textit{Sequence Lengths}} \\
Max prompt length & 512 tokens \\
Max completion length & 3584 tokens \\
Max model length & 4608 tokens \\
\midrule
\multicolumn{2}{l}{\textit{Generation}} \\
Temperature & 0.7 \\
Number of generations ($G$) & 6 \\
\midrule
\multicolumn{2}{l}{\textit{System}} \\
Random seed & 2025 \\
Gradient checkpointing & True \\
Mixed precision & bfloat16 \\
\bottomrule
\end{tabular}}
\caption{Common training hyperparameters across all methods and models.}
\label{tab:common_hyperparams}
\end{table}

\subsection{vLLM Generation Configuration}

For efficient parallel generation during training, we use vLLM \cite{kwon2023efficientmemorymanagementlarge} with the configuration shown in Table~\ref{tab:vllm_config}.

\begin{table}[h]
\centering
\small
\begin{tabular}{lc}
\toprule
\textbf{Parameter} & \textbf{Value} \\
\midrule
vLLM device & 1 Auto-assigned GPU \\
GPU memory utilization & 0.7 \\
Max model length & 4608 tokens \\
Eager mode & Enabled \\
Prefix caching & Enabled \\
Dtype & bfloat16 \\
\bottomrule
\end{tabular}
\caption{vLLM configuration for efficient generation.}
\label{tab:vllm_config}
\end{table}

\subsection{Reward Functions}
Table~\ref{tab:reward_funcs} describes the reward functions used in our experiments.

\begin{table}[h]
\centering
\small
\scalebox{0.9}{
\begin{tabular}{lp{5.5cm}}
\toprule
\textbf{Reward Type} & \textbf{Description} \\
\midrule
Accuracy & Binary reward (1.0 or 0.0) based on whether the extracted answer matches the gold answer \\
\midrule
Format & Measures compliance with the required output format (presence of \texttt{<think>} and \texttt{<answer>} tags, proper \texttt{\textbackslash boxed\{\}} usage) \\
\midrule
Cosine & This is an enhanced variant of the Accuracy Reward. It assigns rewards to model completions by jointly considering solution correctness and completion length, where the length-based scaling follows a cosine schedule. For each completion, both the model output and the reference solution are parsed. Correctness is determined by comparing the parsed representations. The final reward is computed as a cosine function of the completion length normalized by a predefined maximum length, which favors shorter correct responses while imposing stronger penalties on short but incorrect ones \\
\bottomrule
\end{tabular}}
\caption{Description of reward functions used across different methods.}
\label{tab:reward_funcs}
\end{table}

\subsection{Method-Specific Configurations}
Table~\ref{tab:method_configs} compares the specific hyperparameters and configurations for each training method.

\begin{table*}[t]
\centering
\small
\begin{tabular}{lccc}
\toprule
\textbf{Configuration} & \textbf{DAPO} & \textbf{GRPO} & \textbf{DR-GRPO} \\
\midrule
\multicolumn{4}{l}{\textit{Algorithm-specific Parameters}} \\
Clipping bound (lower) $\epsilon_{\text{low}}$ & 0.2 & -- & -- \\
Clipping bound (upper) $\epsilon_{\text{high}}$ & 0.28 & -- & -- \\
KL penalty coefficient $\beta$ & -- & 0.04 & 0.04 \\
Dynamic sampling & Applicable & N/A & N/A \\
Filter reward index & 0 (accuracy) & -- & -- \\
Max generation batches & 10 & -- & -- \\
\midrule
\multicolumn{4}{l}{\textit{Reward Functions and Weights}} \\
Primary reward & Accuracy (1.0) & Format (1.0) & Format (1.0) \\
Secondary reward & Cosine (1.0) & Cosine (2.0) & Cosine (2.0) \\
\bottomrule
\end{tabular}
\caption{Method-specific configurations for DAPO, GRPO, and DR-GRPO.}
\label{tab:method_configs}
\end{table*}

\subsection{System Prompt}

All models use the following system prompt during training:

\begin{quote}
\small
\textit{``A conversation between User and Assistant. The user asks a question, and the Assistant solves it. The assistant first thinks about the reasoning process in the mind and then provides the user with the answer, and put your final answer within \textbackslash boxed\{\} . The reasoning process and answer are enclosed within <think> </think> and <answer> </answer> tags, respectively, i.e., <think> reasoning process here </think> <answer> answer here </answer>. Note that respond by English, NOT use other languages.''}
\end{quote}

All models use the following system prompt during evaluation:

\begin{quote}
\small
\textit{``Solve the following math problem efficiently and clearly.  The last line of your response should be of the following format: 'Therefore, the final answer is: \$\textbackslash\textbackslash boxed\{\{ANSWER\}\}\$. I hope it is correct' (without quotes) where ANSWER is just the final number or expression that solves the problem. Think step by step before answering.
''}
\end{quote}

\subsection{Computational Resources}
All training experiments were conducted on 2xNVIDIA H100 80GB GPUs. All evaluation experiments were conducted on 1xNVIDIA A100 40GB GPU.

\section{Adaptive $\lambda$ Design Choice}
The design is motivated by two principles grounded in the MMR literature:
\begin{enumerate}[noitemsep,topsep=0pt]
    \item Lower bound of 0.5: The original MMR paper \cite{10.1145/290941.291025} establishes $\lambda = 0.7$ as the default, reflecting that relevance should generally dominate over diversity. Our sigmoid maps to $[0.5, 1)$, ensuring that even in the most diversity-favoring regime (low reward variance), reward quality always receives at least equal weight as diversity — quality is never subordinated.
    \item Sigmoid function: The sigmoid naturally maps the unbounded reward standard deviation to a bounded range with smooth, monotonic behavior. When $std(r)$ is small (rewards concentrated), $\lambda$ approaches 0.5, increasing the diversity emphasis; when $std(r)$ is large (rewards spread), $\lambda$ approaches 1, reverting to reward-dominated selection.
\end{enumerate}

\section{Additional Results}
\label{app:additional_results}

\subsection{Training Convergence for 8B and 1.5B Models}
\label{app:convergence_7b_8b}

\begin{figure*}[t]
    \centering
    \begin{subfigure}[t]{\textwidth}
        \centering
        \includegraphics[width=\linewidth]{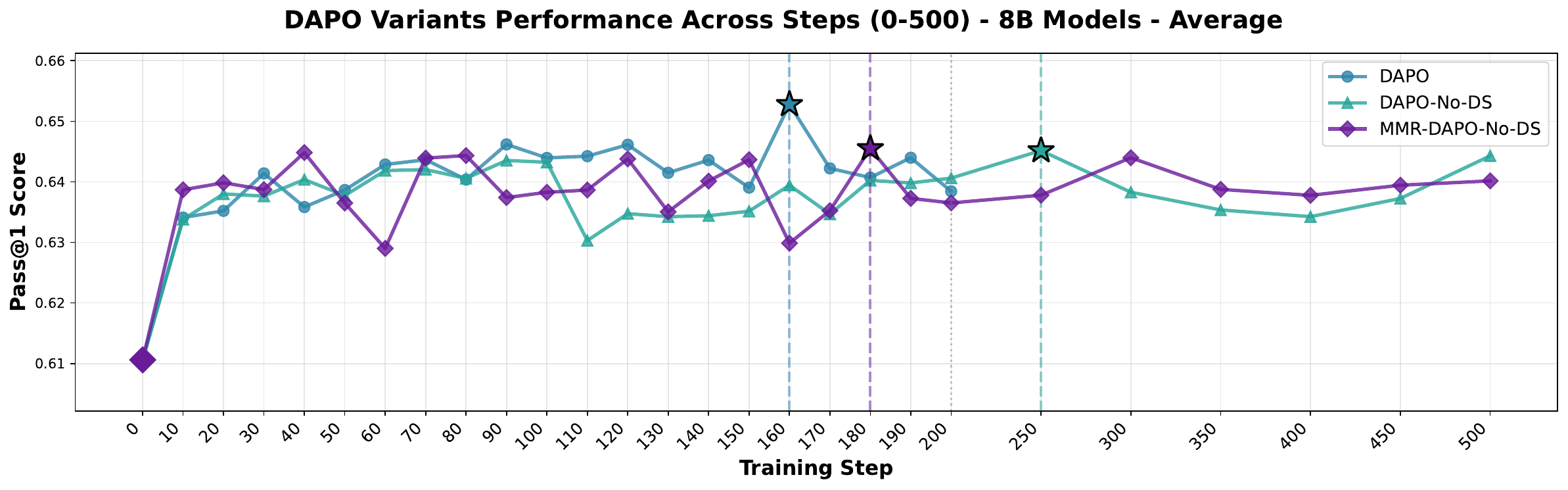}
    \end{subfigure}

    \begin{subfigure}[t]{0.48\textwidth}
        \centering
        \includegraphics[width=\linewidth]{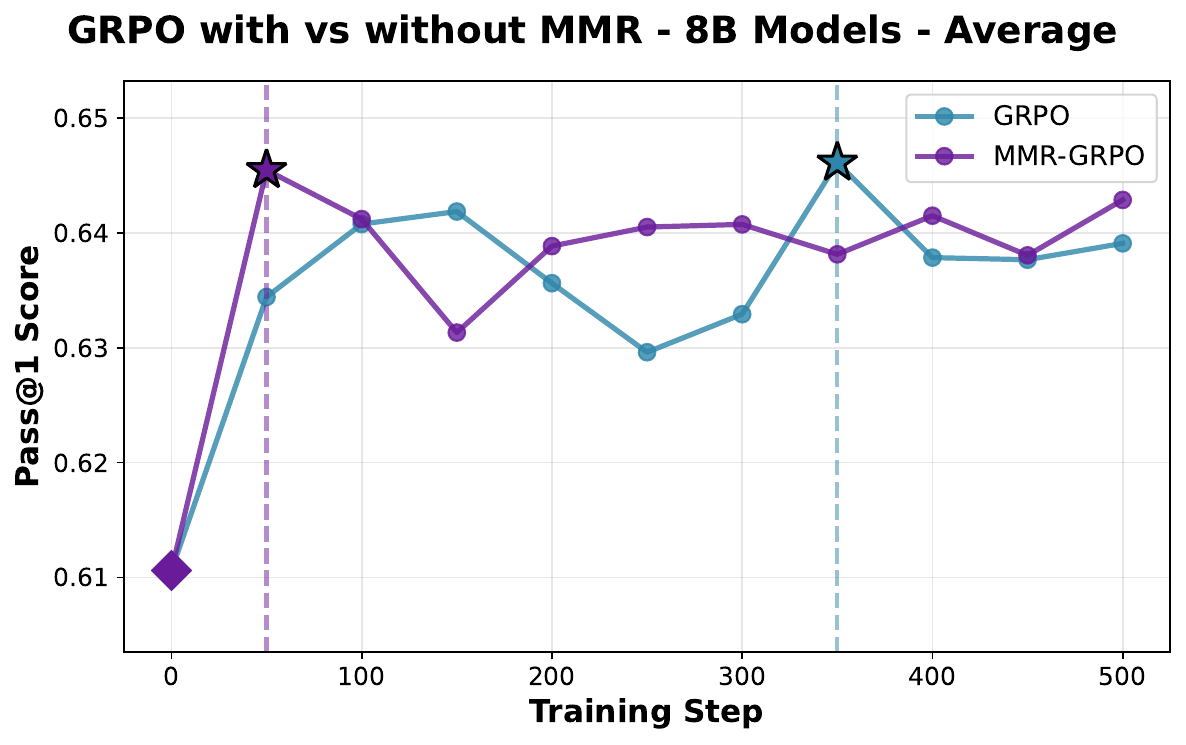}
    \end{subfigure}
    \hfill
    \begin{subfigure}[t]{0.48\textwidth}
        \centering
        \includegraphics[width=\linewidth]{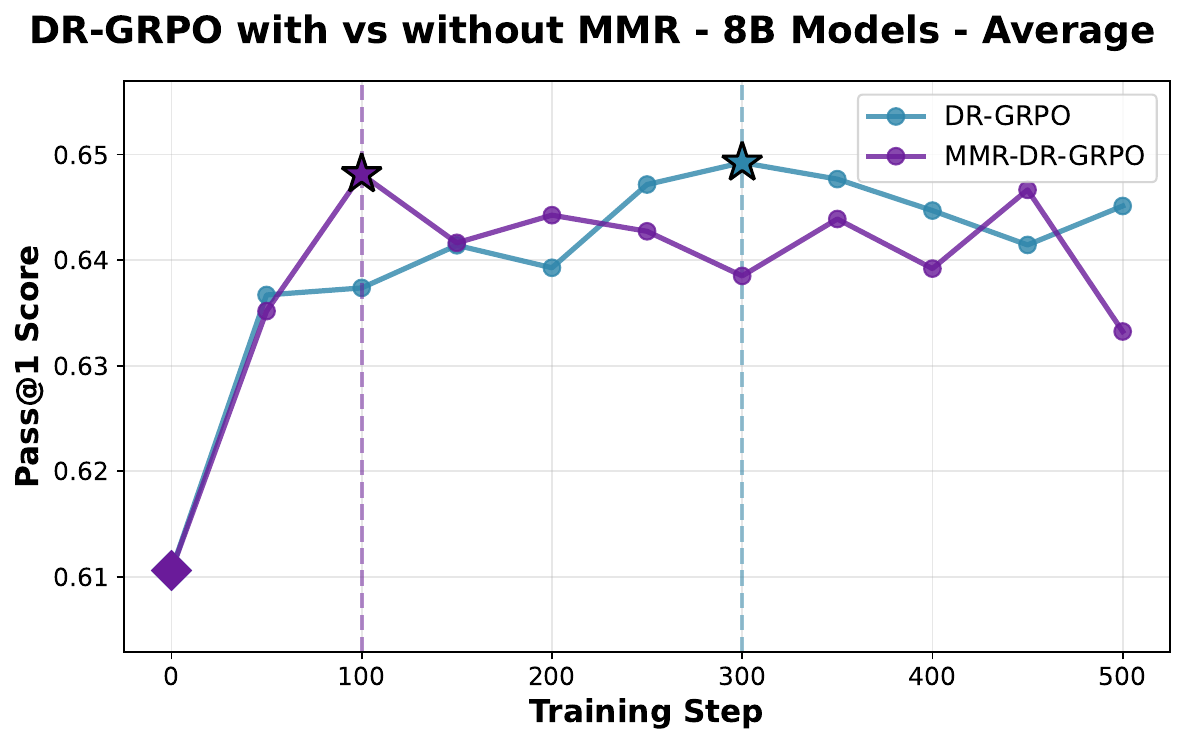}
    \end{subfigure}
    \vspace{-0.2cm}
    \caption{Performance across training steps for 8B models across all three training methods (DAPO, GRPO, DR-GRPO). MMR variants consistently achieve faster convergence and reach peak performance with fewer training steps.}
    \label{fig:8B}
\end{figure*}

\begin{figure*}[t]
    \centering
    \begin{subfigure}[t]{\textwidth}
        \centering
        \includegraphics[width=\linewidth]{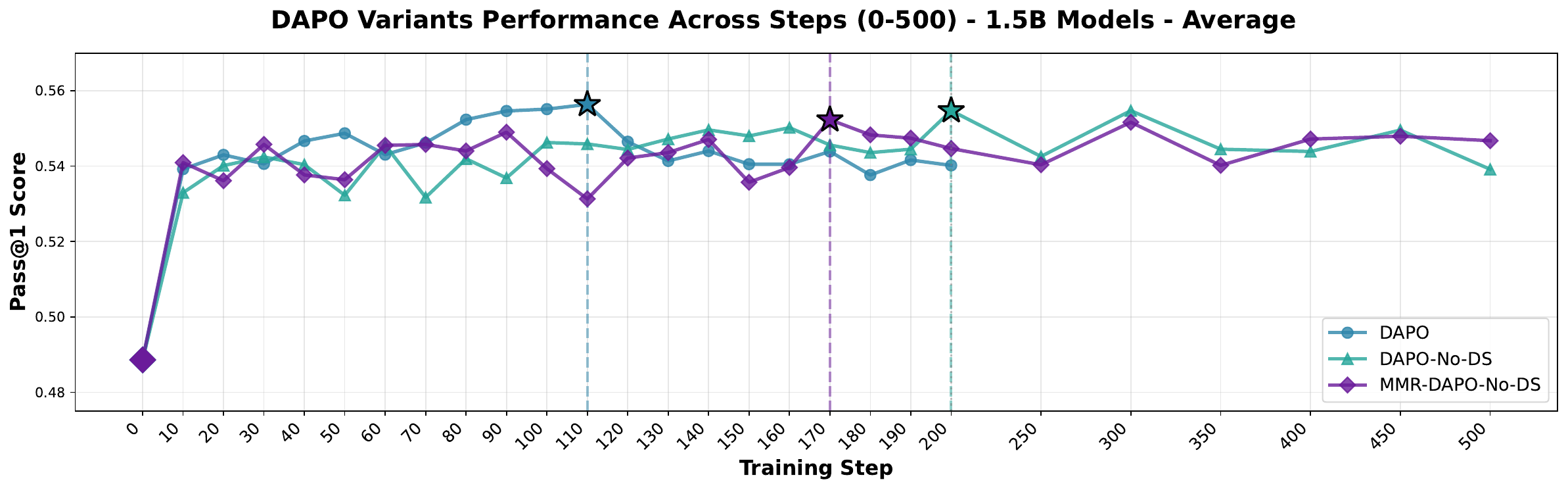}
    \end{subfigure}

    \begin{subfigure}[t]{0.48\textwidth}
        \centering
        \includegraphics[width=\linewidth]{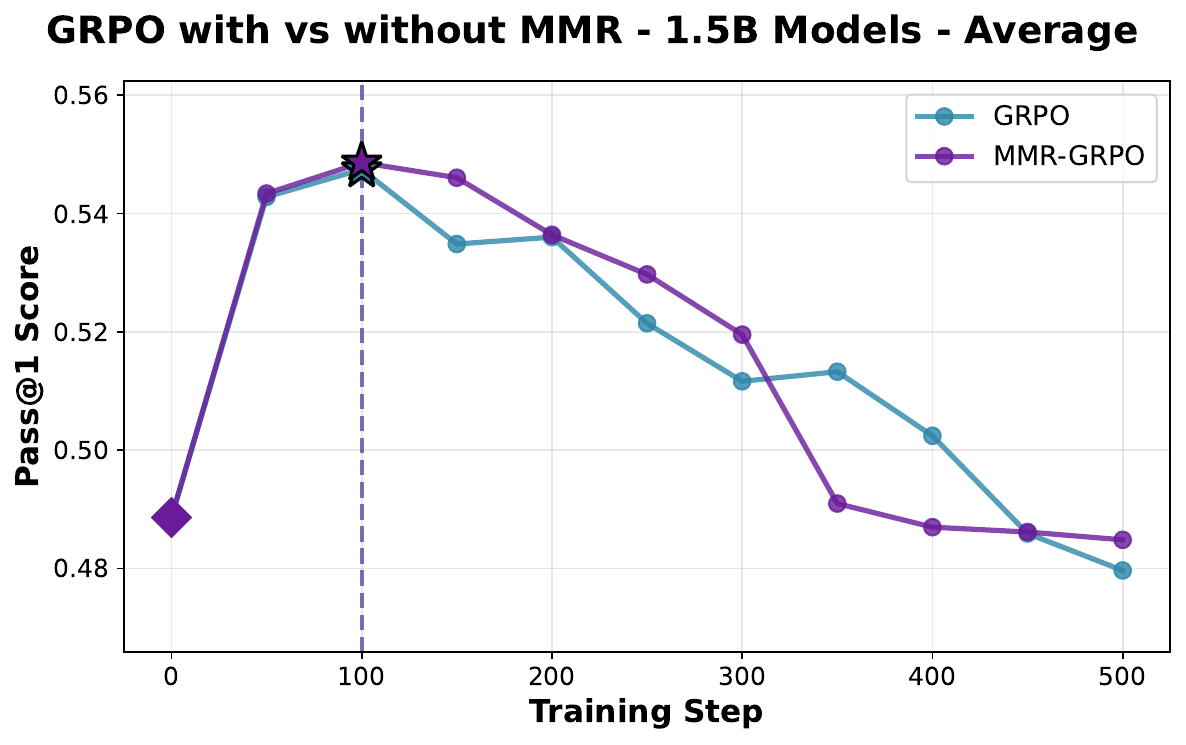}
    \end{subfigure}
    \hfill
    \begin{subfigure}[t]{0.48\textwidth}
        \centering
        \includegraphics[width=\linewidth]{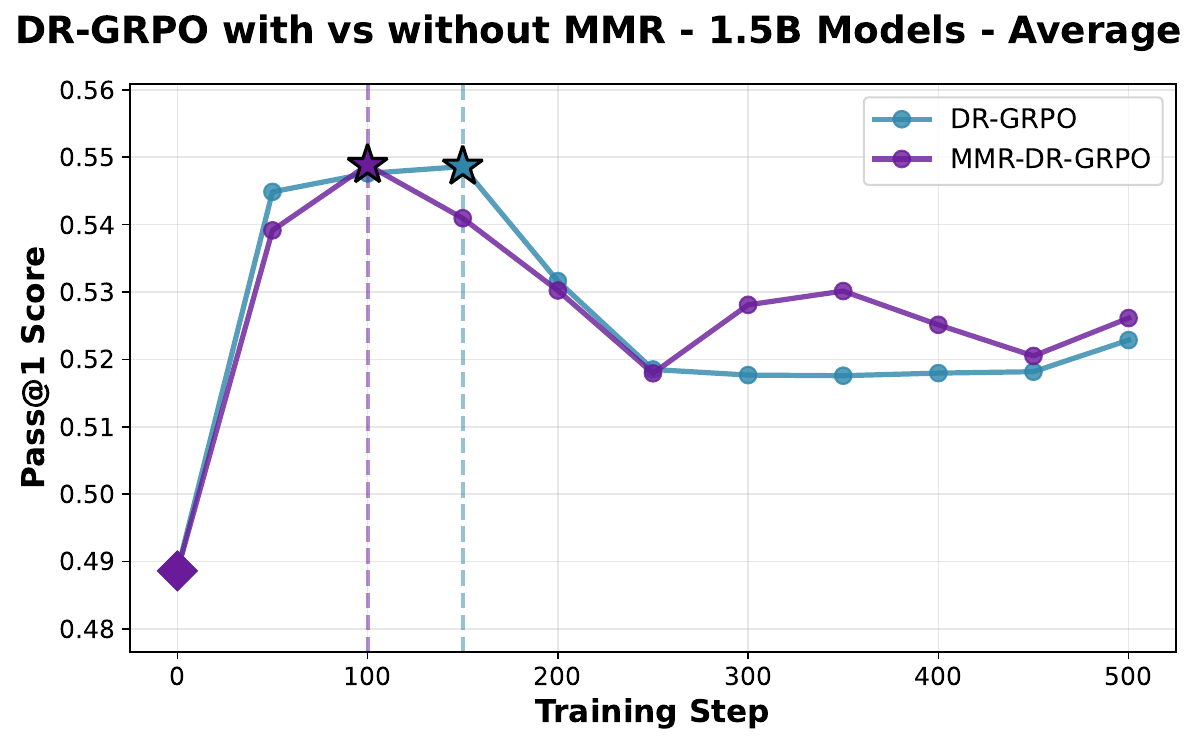}
    \end{subfigure}
    \vspace{-0.2cm}
    \caption{Performance across training steps for 1.5B models across all three training methods (DAPO, GRPO, DR-GRPO). MMR variants consistently achieve faster convergence and reach peak performance with fewer training steps.}
    \vspace{-0.2cm}
    \label{fig:1.5B}
\end{figure*}

Figures~\ref{fig:8B} and~\ref{fig:1.5B} present the training convergence patterns for 8B and 1.5B models, respectively, complementing the 7B results shown in the main text (Figure~\ref{fig:7B}). The convergence patterns observed in these models are consistent with those in the 7B model.

For 8B models (Figure~\ref{fig:8B}), the efficiency gains are even more dramatic. MMR-GRPO achieves peak performance at step 50 compared to step 350 for vanilla GRPO (86\% reduction), representing the largest step reduction observed across all configurations. MMR-DR-GRPO peaks at step 100 versus step 300 for vanilla DR-GRPO (67\% reduction), while MMR-DAPO-No-DS requires 180 steps compared to 250 for DAPO-No-DS (28\% reduction). These results suggest that the benefits of diversity-aware training may scale favorably with model capacity, as larger models can better exploit the informative signals from diverse completions.

For 1.5B models (Figure~\ref{fig:1.5B}), MMR-GRPO reaches peak performance at step 100 compared to step 100 for vanilla GRPO, while MMR-DR-GRPO achieves peak performance at step 150 versus step 100 for vanilla DR-GRPO. The DAPO comparison shows that MMR-DAPO-No-DS peaks at step 170, representing a 15\% reduction from the 200 steps required by DAPO-No-DS.

Across both model scales, the performance curves demonstrate that MMR variants not only converge faster but also maintain comparable peak performance compared to their baseline counterparts. The consistency of these patterns across different model sizes (1.5B, 7B, 8B), training methods (GRPO, DR-GRPO, DAPO), and benchmarks provides strong evidence for the generality and robustness of the MMR-based reward reweighting approach.

\subsection{Exploration Ability: Pass@k Analysis for GRPO and DR-GRPO}
\label{app:passk_grpo_drgrpo}

\begin{figure*}[t]
    \centering
    \begin{subfigure}{0.32\linewidth}
        \centering
        \includegraphics[width=\linewidth]{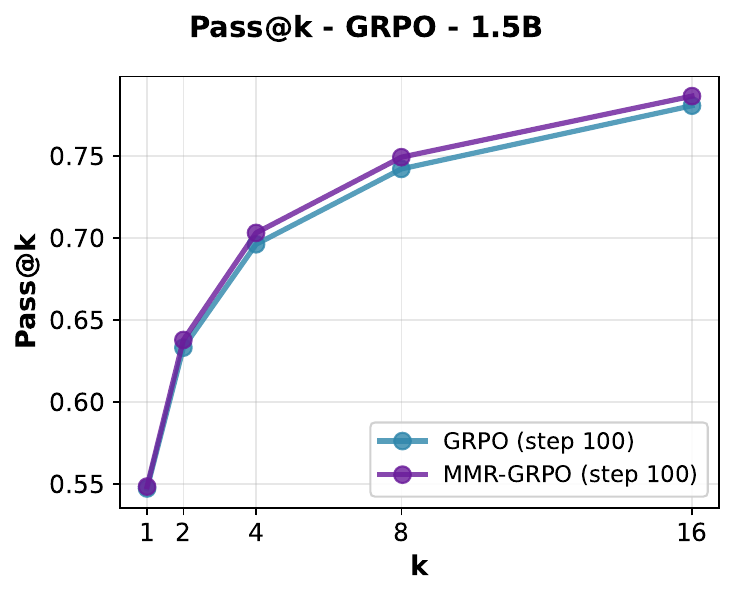}
    \end{subfigure}
    \hfill
    \begin{subfigure}{0.32\linewidth}
        \centering
        \includegraphics[width=\linewidth]{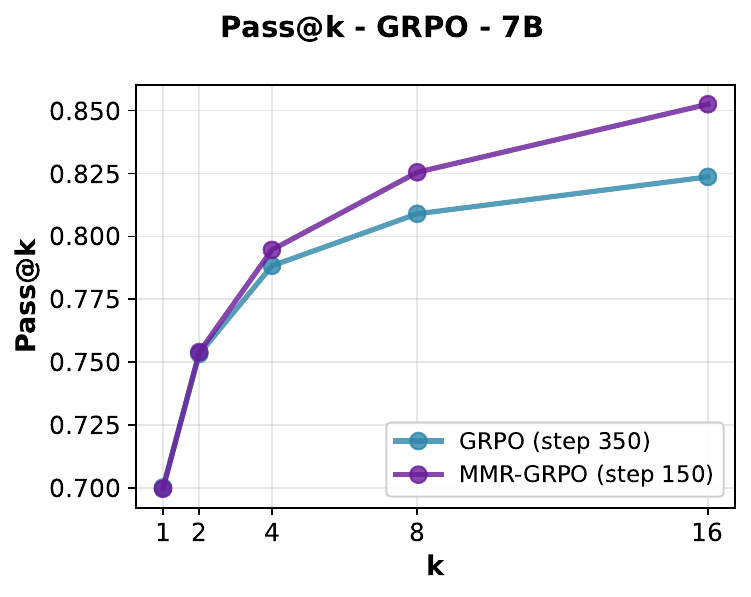}
    \end{subfigure}
    \hfill
    \begin{subfigure}{0.32\linewidth}
        \centering
        \includegraphics[width=\linewidth]{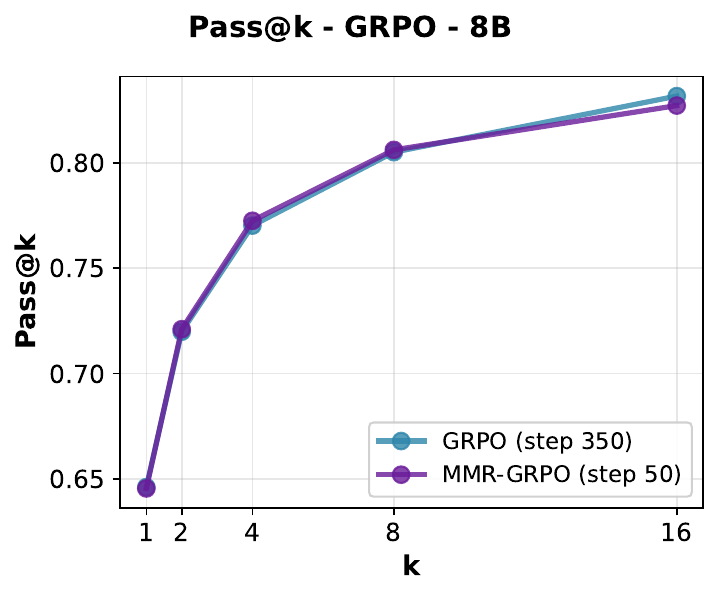}
    \end{subfigure}
    \vspace{-0.2cm}
    \caption{Pass@k curves for GRPO methods across three model scales (1.5B, 7B, 8B). MMR-GRPO maintains nearly identical pass@k performance to vanilla GRPO across all k values.}
    \label{fig:pass_at_k_grpo}
\end{figure*}

\begin{figure*}[t]
    \centering
    \begin{subfigure}{0.32\linewidth}
        \centering
        \includegraphics[width=\linewidth]{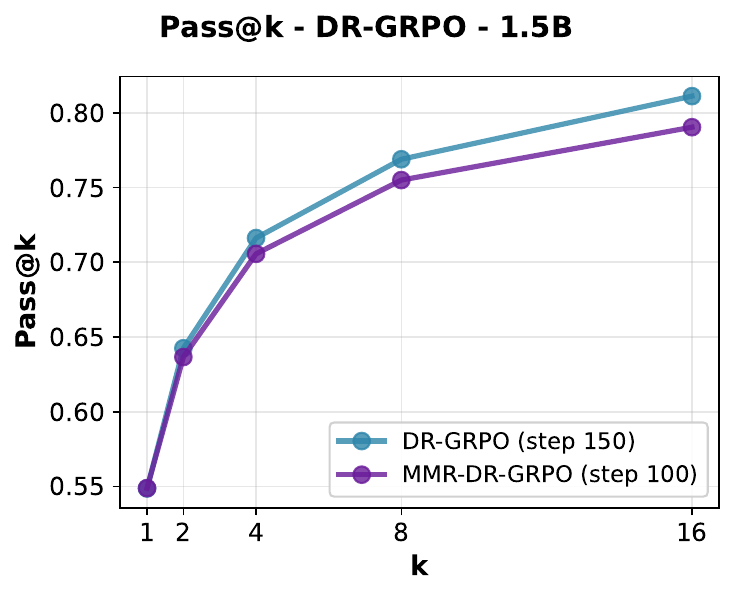}
    \end{subfigure}
    \hfill
    \begin{subfigure}{0.32\linewidth}
        \centering
        \includegraphics[width=\linewidth]{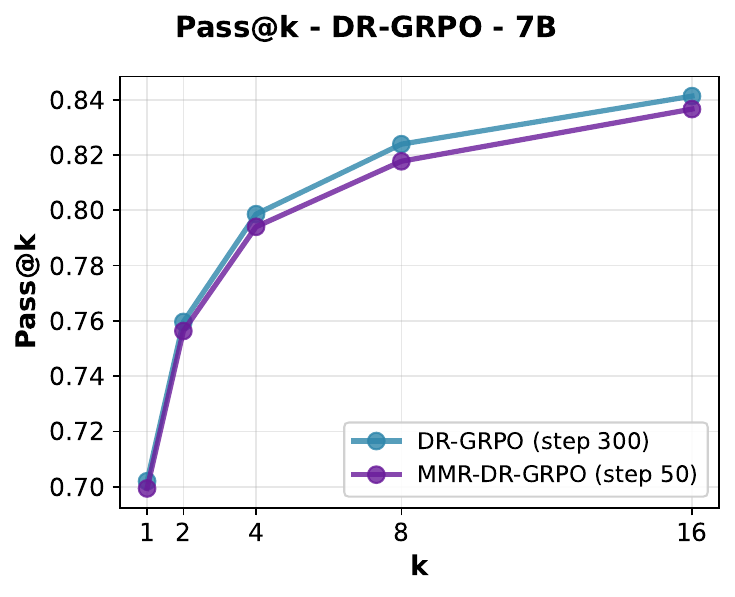}
    \end{subfigure}
    \hfill
    \begin{subfigure}{0.32\linewidth}
        \centering
        \includegraphics[width=\linewidth]{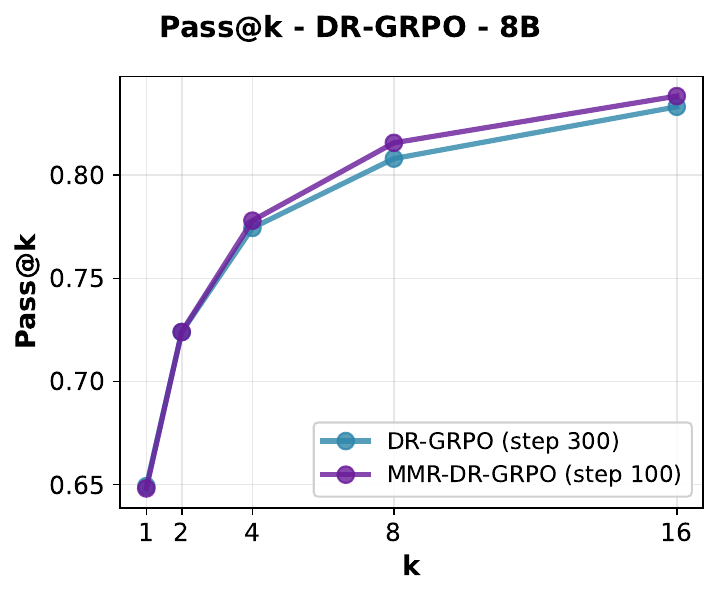}
    \end{subfigure}
    \vspace{-0.2cm}
    \caption{Pass@k curves for DR-GRPO methods across three model scales (1.5B, 7B, 8B). MMR-DR-GRPO shows nearly indistinguishable pass@k performance from vanilla DR-GRPO.}
    \label{fig:pass_at_k_drgrpo}
\end{figure*}

Figures~\ref{fig:pass_at_k_grpo} and~\ref{fig:pass_at_k_drgrpo} present pass@k curves for GRPO and DR-GRPO methods across all three model scales, complementing the DAPO pass@k analysis shown in the main text (Figure~\ref{fig:pass_at_k_dapo}).

For GRPO methods (Figure~\ref{fig:pass_at_k_grpo}), MMR-GRPO and vanilla GRPO exhibit nearly identical pass@k curves across all model sizes and k values. At k=1 (equivalent to majority voting), both methods achieve similar performance. As k increases, the curves remain tightly coupled, with both methods reaching comparable coverage at k=16 (approximately 0.78 for 1.5B, 0.82 for 7B, and 0.82 for 8B). The overlapping curves indicate that MMR-trained GRPO models retain the same diversity of solution strategies as vanilla GRPO models at inference time.

Similarly, for DR-GRPO methods (Figure~\ref{fig:pass_at_k_drgrpo}), MMR-DR-GRPO and vanilla DR-GRPO show nearly indistinguishable pass@k performance across all configurations. The curves overlap across the entire range of k values (1-16) for all three model scales, with final coverage at k=16 being virtually identical between MMR and non-MMR variants (approximately 0.80 for 1.5B, 0.84 for 7B, and 0.83 for 8B).

These results, combined with the DAPO pass@k analysis in the main text, provide comprehensive evidence that MMR-based reward reweighting does not compromise the trained model's ability to explore diverse solution strategies during inference. The consistency of this finding across all three training methods (GRPO, DR-GRPO, DAPO) and all three model scales (1.5B, 7B, 8B) demonstrates that diversity-aware training enhances training efficiency without restricting the model's representational flexibility or exploration capabilities.

\section{Performance Across Training Steps on Each Dataset}
\label{app:performance}
This section provides a comprehensive analysis of model performance evolution across training steps on all evaluation benchmarks. We present results for three model sizes (1.5B, 7B, 8B) across five mathematical reasoning benchmarks: MATH-500, AIME 2024, Minerva Math, AMC 2023, OlympiadBench, and their average performance.

\subsection{GRPO and DR-GRPO Performance Across Training Steps on Each Dataset}
\begin{figure*}[t]
    \centering
    \includegraphics[width=0.95\textwidth]{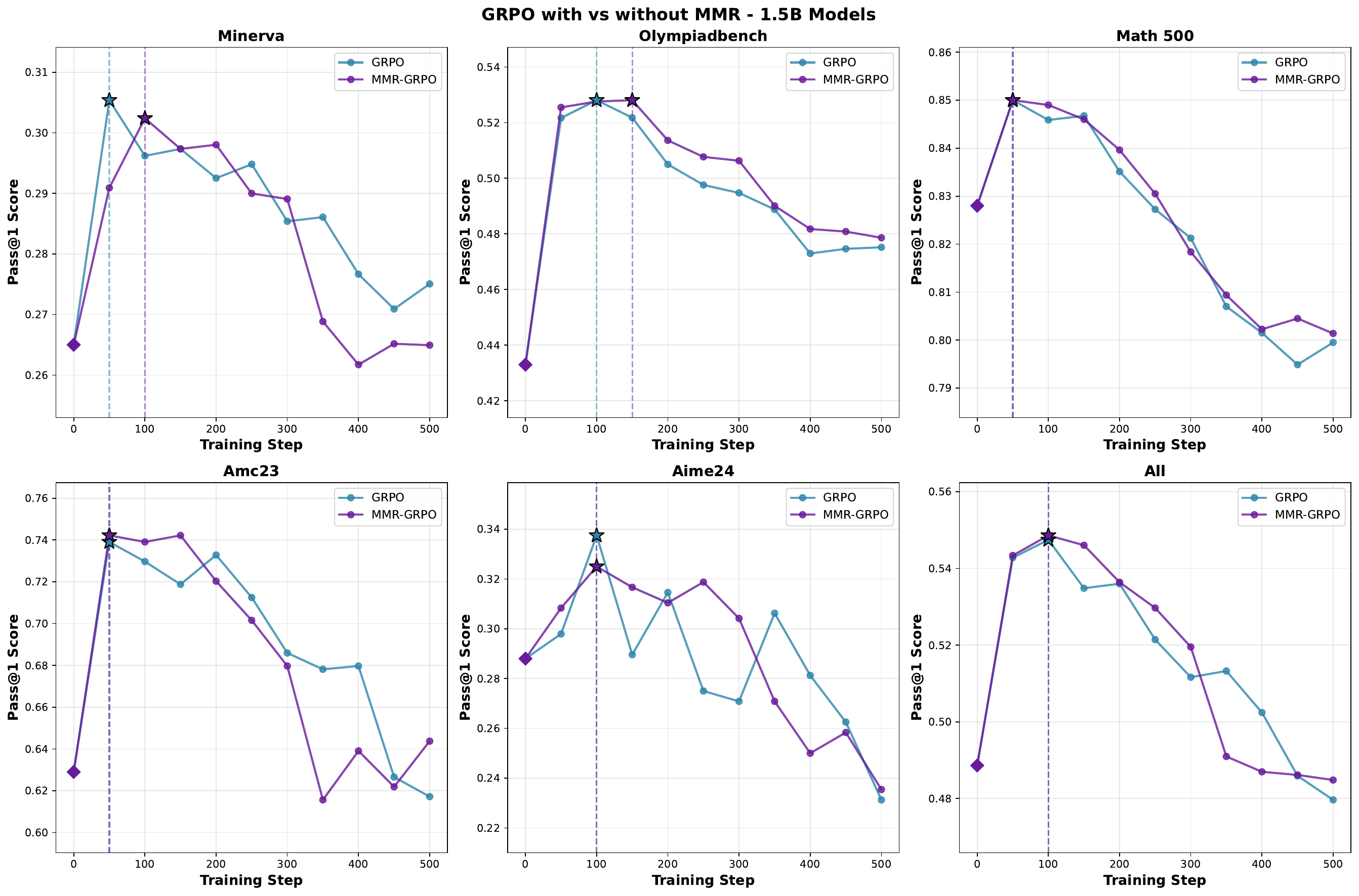}
    \caption{Pass@1 performance across training steps for GRPO with and without MMR reweighting on 1.5B models. Each subplot shows results for a specific benchmark. Diamond markers indicate baseline performance (step 0), star markers show peak performance, and dashed vertical lines indicate the optimal training step for each configuration.}
    \label{fig:grpo_1.5b}
\end{figure*}

\begin{figure*}[t]
    \centering
    \includegraphics[width=0.95\textwidth]{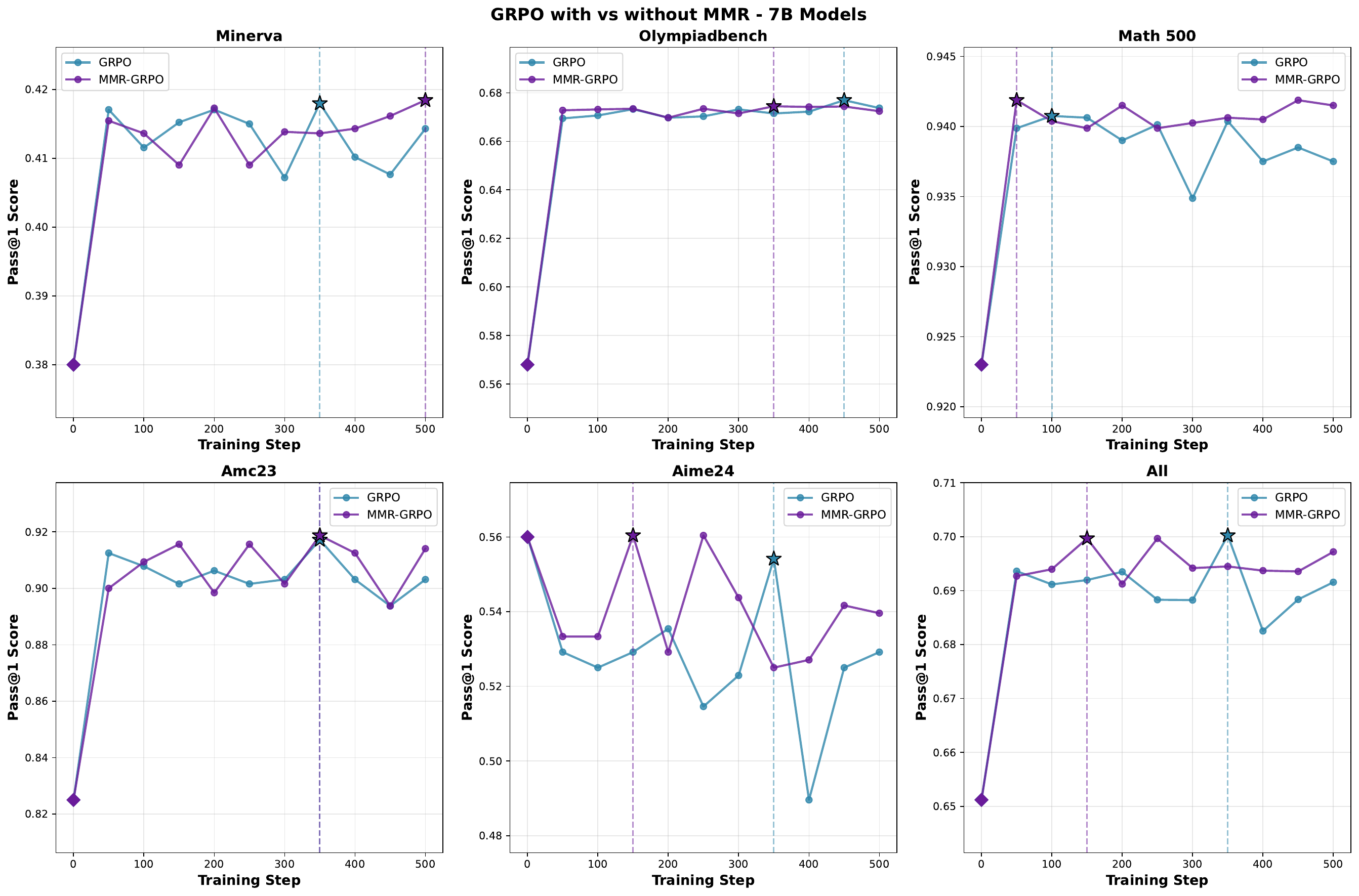}
    \caption{Pass@1 performance across training steps for GRPO with and without MMR reweighting on 7B models. Each subplot shows results for a specific benchmark. Diamond markers indicate baseline performance (step 0), star markers show peak performance, and dashed vertical lines indicate the optimal training step for each configuration.}
    \label{fig:grpo_7b}
\end{figure*}

\begin{figure*}[t]
    \centering
    \includegraphics[width=0.95\textwidth]{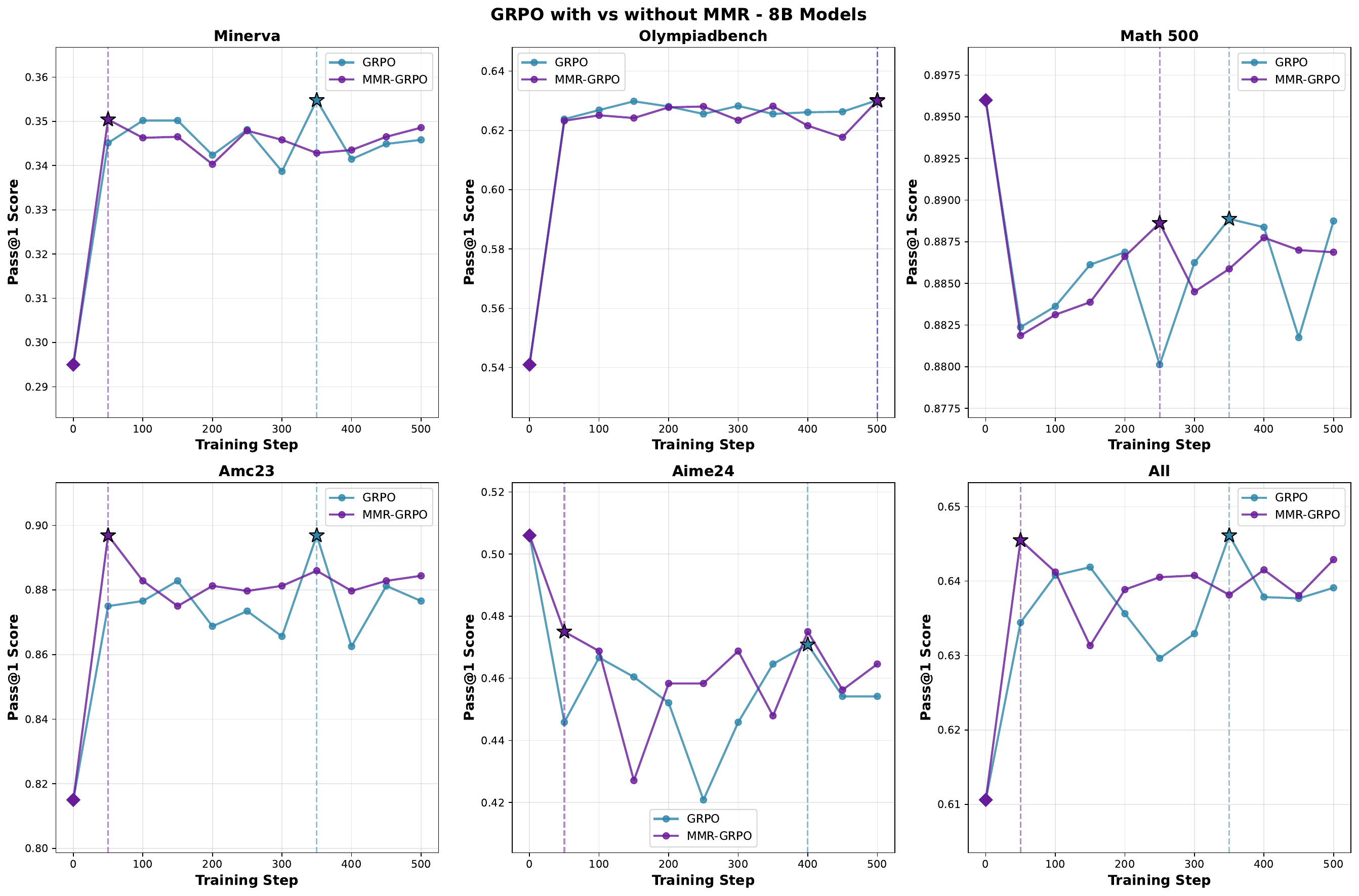}
    \caption{Pass@1 performance across training steps for GRPO with and without MMR reweighting on 8B models. Each subplot shows results for a specific benchmark. Diamond markers indicate baseline performance (step 0), star markers show peak performance, and dashed vertical lines indicate the optimal training step for each configuration.}
    \label{fig:grpo_8b}
\end{figure*}

\begin{figure*}[t]
    \centering
    \includegraphics[width=0.95\textwidth]{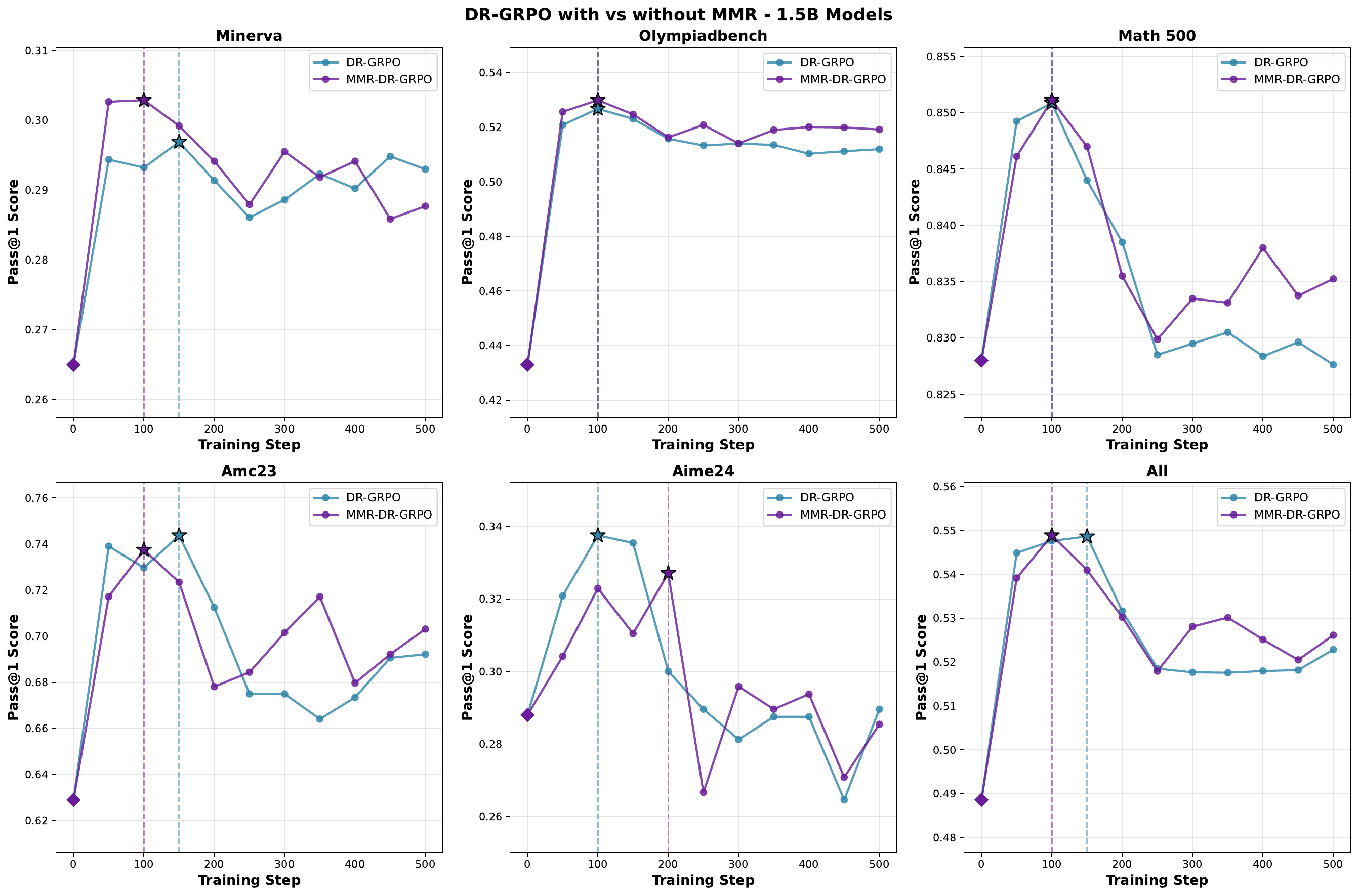}
    \caption{Pass@1 performance across training steps for DR-GRPO with and without MMR reweighting on 1.5B models. Each subplot shows results for a specific benchmark. Diamond markers indicate baseline performance (step 0), star markers show peak performance, and dashed vertical lines indicate the optimal training step for each configuration.}
    \label{fig:drgrpo_1.5b}
\end{figure*}

\begin{figure*}[t]
    \centering
    \includegraphics[width=0.95\textwidth]{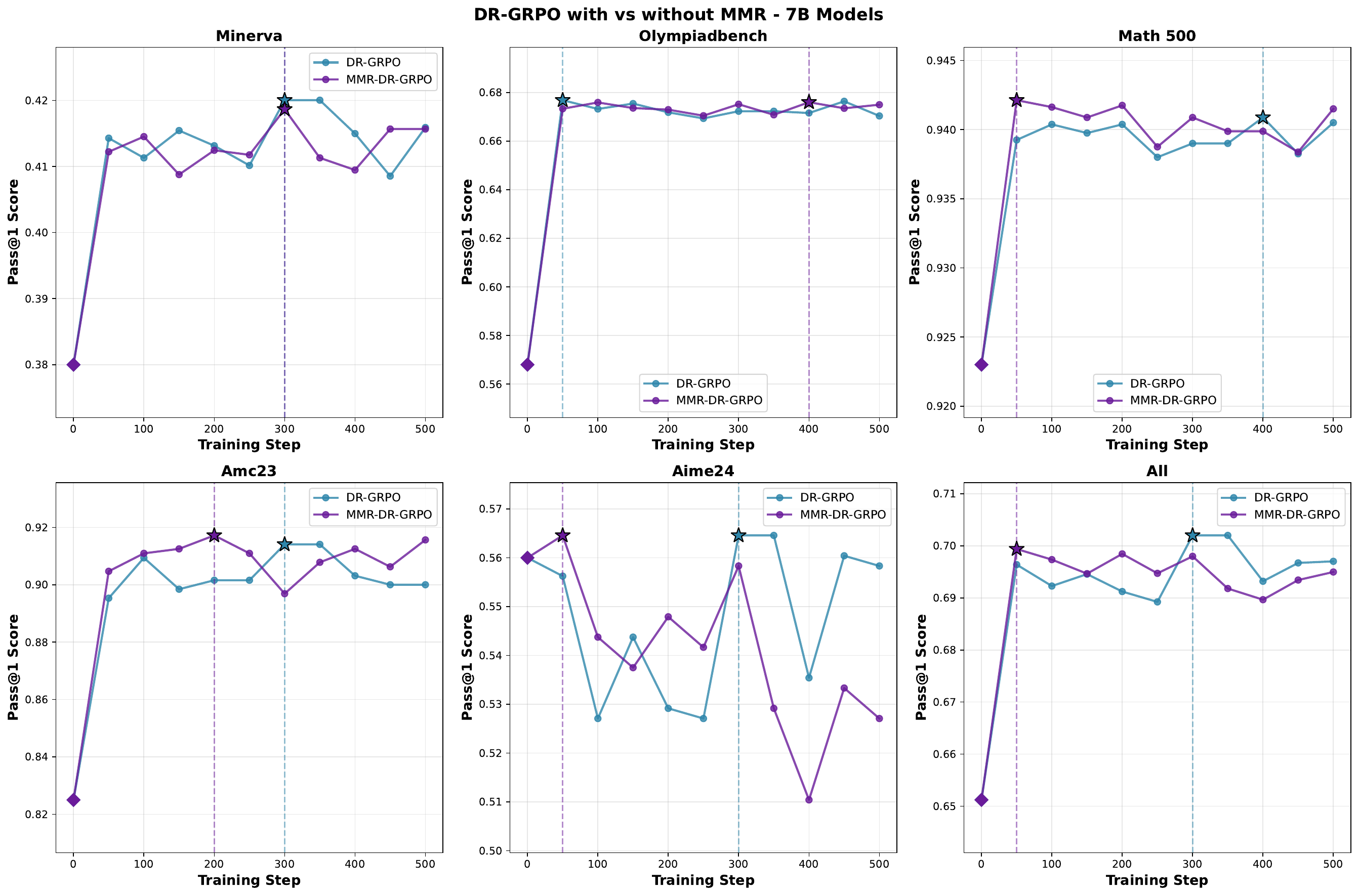}
    \caption{Pass@1 performance across training steps for DR-GRPO with and without MMR reweighting on 7B models. Each subplot shows results for a specific benchmark. Diamond markers indicate baseline performance (step 0), star markers show peak performance, and dashed vertical lines indicate the optimal training step for each configuration.}
    \label{fig:drgrpo_7b}
\end{figure*}

\begin{figure*}[t]
    \centering
    \includegraphics[width=0.95\textwidth]{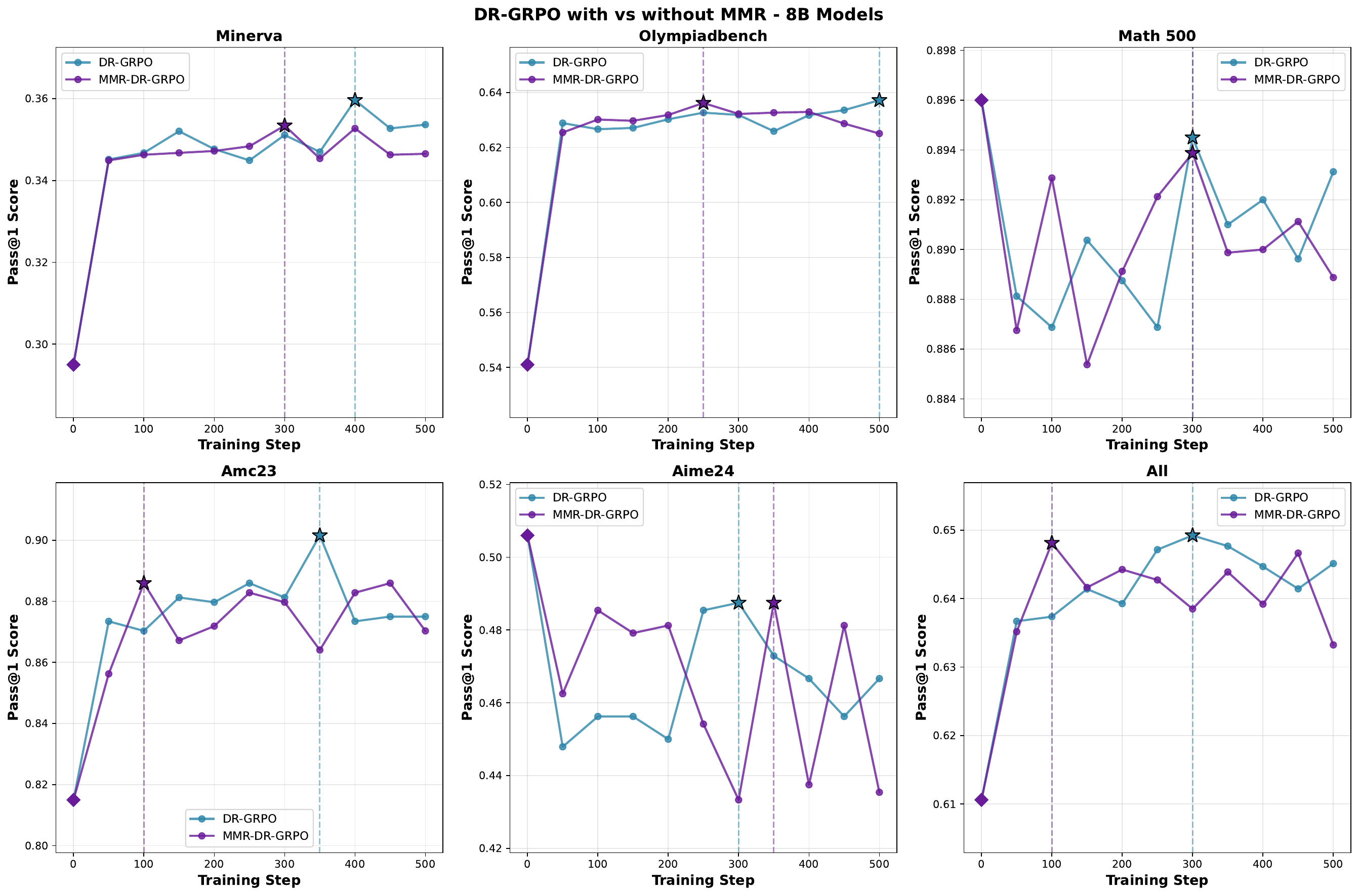}
    \caption{Pass@1 performance across training steps for DR-GRPO with and without MMR reweighting on 8B models. Each subplot shows results for a specific benchmark. Diamond markers indicate baseline performance (step 0), star markers show peak performance, and dashed vertical lines indicate the optimal training step for each configuration.}
    \label{fig:drgrpo_8b}
\end{figure*}

Figures~\ref{fig:grpo_1.5b}, \ref{fig:grpo_7b}, and \ref{fig:grpo_8b} illustrate the pass@1 performance evolution for GRPO method with and without MMR reweighting across 500 training steps.
Figures~\ref{fig:drgrpo_1.5b}, \ref{fig:drgrpo_7b}, and \ref{fig:drgrpo_8b} illustrate the pass@1 performance evolution for DR-GRPO method with and without MMR reweighting across 500 training steps. Each figure contains six subplots corresponding to individual benchmarks and their average.

\subsection{DAPO Performance Across Training Steps on Each Dataset}
\begin{figure*}[t]
    \centering
    \includegraphics[width=0.95\textwidth]{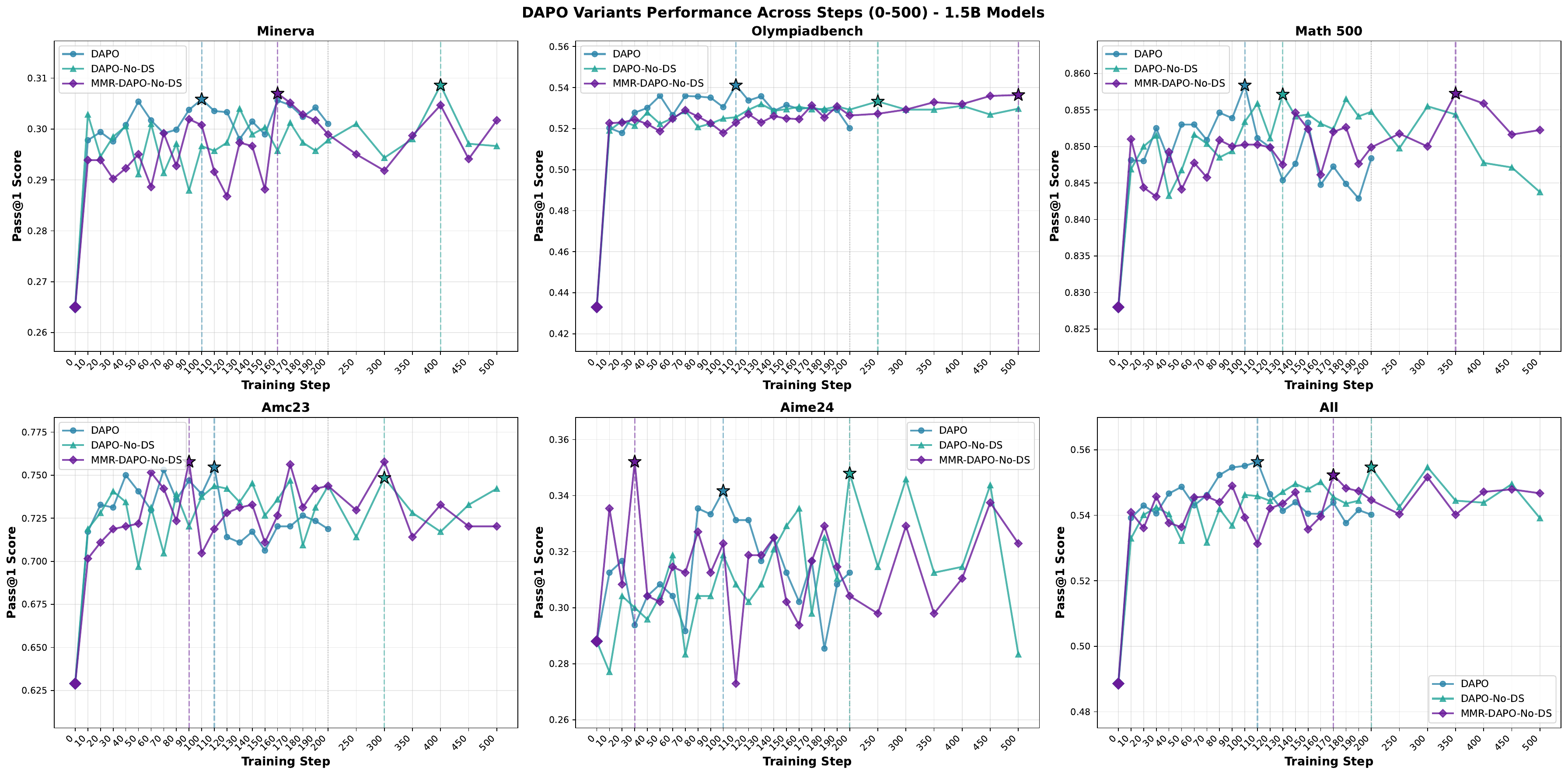}
    \caption{Pass@1 performance across extended training steps (0-500) for DAPO variants on 1.5B models. The x-axis uses non-linear scaling: steps 0-200 occupy 60\% of the space, steps 200-500 occupy 40\%. The gray dotted line marks step 200. Star markers show peak performance.}
    \label{fig:dapo_ext_1.5b}
\end{figure*}

\begin{figure*}[t]
    \centering
    \includegraphics[width=0.95\textwidth]{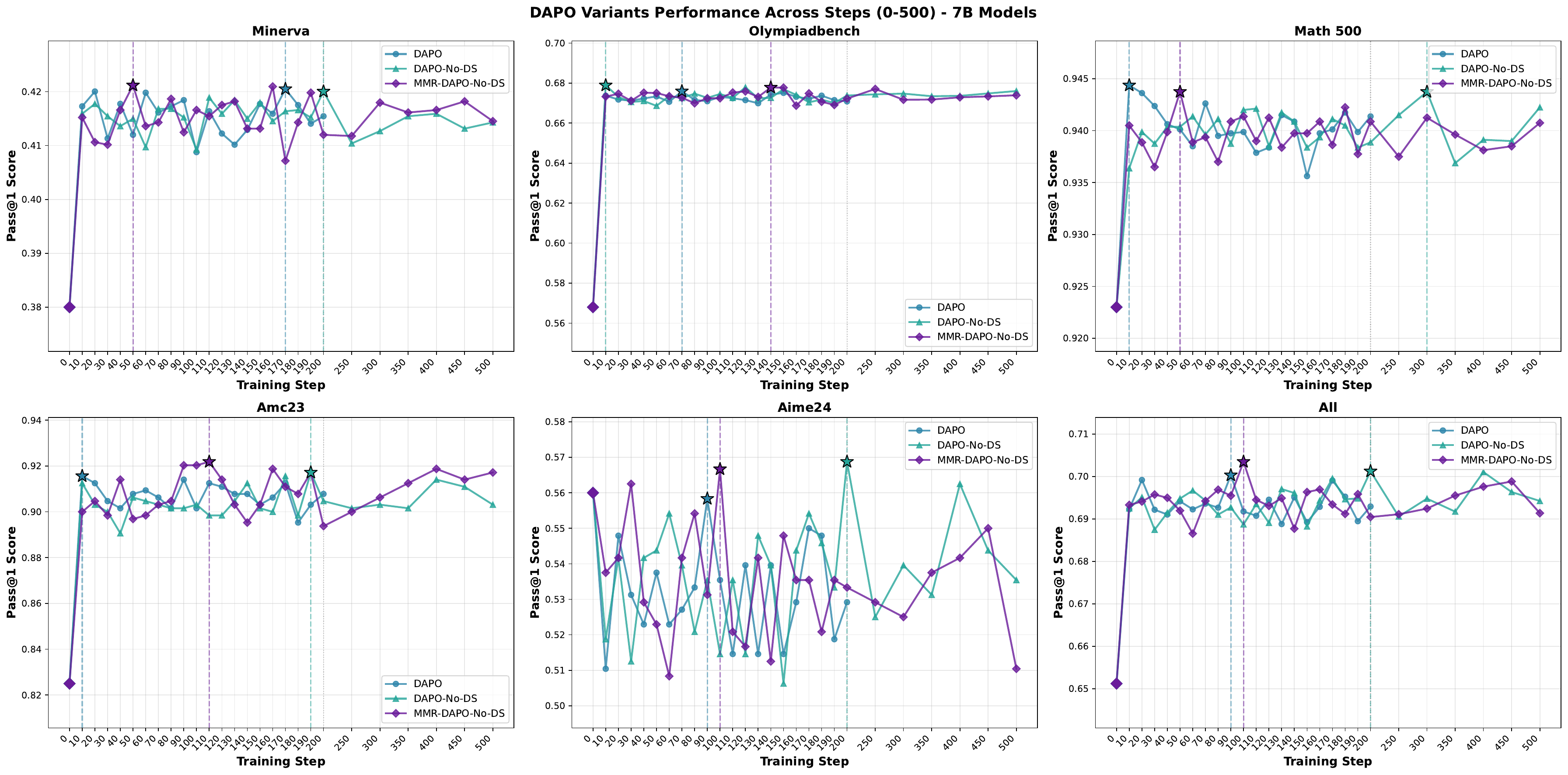}
    \caption{Pass@1 performance across extended training steps (0-500) for DAPO variants on 7B models. The x-axis uses non-linear scaling: steps 0-200 occupy 60\% of the space, steps 200-500 occupy 40\%. The gray dotted line marks step 200. Star markers show peak performance.}
    \label{fig:dapo_ext_7b}
\end{figure*}

\begin{figure*}[t]
    \centering
    \includegraphics[width=0.95\textwidth]{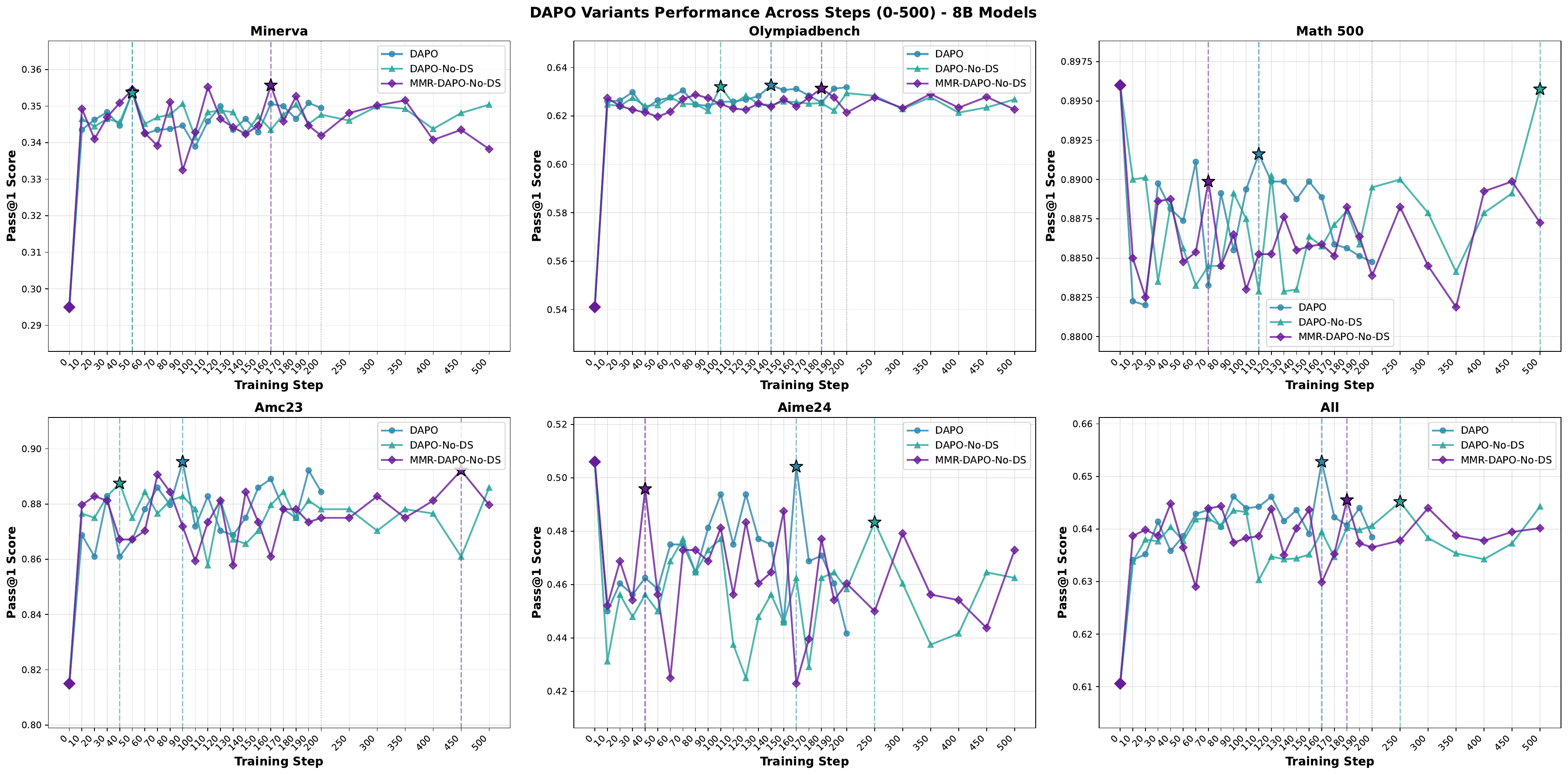}
    \caption{Pass@1 performance across extended training steps (0-500) for DAPO variants on 8B models. The x-axis uses non-linear scaling: steps 0-200 occupy 60\% of the space, steps 200-500 occupy 40\%. The gray dotted line marks step 200. Star markers show peak performance.}
    \label{fig:dapo_ext_8b}
\end{figure*}

Figures~\ref{fig:dapo_ext_1.5b}, \ref{fig:dapo_ext_7b}, and \ref{fig:dapo_ext_8b} present DAPO performance across an extended training range (0-500 steps) using a non-linear x-axis scale. The x-axis allocates 60\% of space to steps 0-200 (critical training phase) and 40\% to steps 200-500 (overfitting phase). A gray dotted vertical line at step 200 marks this transition.

\end{document}